\pdfoutput=1

\documentclass[11pt]{article}

\usepackage[preprint]{acl}

\usepackage{times}
\usepackage{latexsym}

\usepackage[T1]{fontenc}

\usepackage[utf8]{inputenc}

\usepackage{microtype}

\usepackage{inconsolata}

\usepackage{times}
\usepackage{soul}
\usepackage{url}
\usepackage[utf8]{inputenc}
\usepackage{caption}
\usepackage{tikz}
\usepackage{amsmath,amsfonts, bm}

\usepackage[ruled,linesnumbered,vlined]{algorithm2e}
\usepackage{algpseudocode}

\usepackage{graphicx}
\usepackage{textcomp}
\usepackage{booktabs} 
\usepackage{siunitx}
\usepackage{xspace,url}

\usepackage{color}
\usepackage{colortbl}
\usepackage{multirow}
\usepackage{multicol}
\usepackage{makecell}
\usepackage{tabularx}
\usepackage{listings}
\usepackage{graphicx}
\usepackage{caption}
\usepackage{enumitem}
\usepackage{adjustbox}
\usepackage{subcaption}

\usepackage{filecontents}
\usepackage[colorinlistoftodos]{todonotes}

\definecolor{airforceblue}{rgb}{0.36, 0.54, 0.66}
\newcommand{\name}{SwapMoE\xspace}
\newcommand{\core}{Virtual Experts\xspace}

\newcommand{\ie}{\textit{i}.\textit{e}.~}
\newcommand{\eg}{\textit{e}.\textit{g}.~}

\newcommand{\update}[1]{{#1}}

\title{SwapMoE: Serving Off-the-shelf MoE-based Large Language Models with Tunable Memory Budget}

\author{
  Rui Kong$^{1, 2}$\thanks{\ Work was done while the author was interning at Institute for AI Industry Research (AIR), Tsinghua University.} \quad Yuanchun Li$^{2, 3}$\thanks{\ Corresponding authors: Yuanchun Li, Linghe Kong.} \quad Qingtian Feng$^{2, 4}$\footnotemark[1] \quad Weijun Wang$^{2}$ \\ \textbf{Xiaozhou Ye$^{5}$ \quad Ye Ouyang$^{5}$ \quad Linghe Kong$^{1}$\footnotemark[2] \quad Yunxin Liu$^{2, 3}$}
\\
$^{1}$Shanghai Jiao Tong University \\
$^{2}$Institute for AI Industry Research (AIR), Tsinghua University \\
$^{3}$Shanghai Artificial Intelligence Laboratory\\
$^{4}$National University of Singapore \\
$^{5}$AsiaInfo Technologies (China), Inc.
}

\newcommand{\correspondingfootnote}{
    \let\oldthefootnote=\thefootnote
    \renewcommand{\thefootnote}{}
    \footnotetext{$\star$ Authors equally contributed.}
    \footnotetext{\update{$\sharp$ This work was done during their internship at NAVER AI Lab.}}
    \footnotetext{\update{Email to:
    \{hwaran.lee, jungwoo.ha\}@navercorp.com, seokhee.hong@vision.snu.ac.kr
    }}
    \let\thefootnote=\oldthefootnote
}

\begin{document}
\maketitle
\begin{abstract} 
  
Mixture of experts (MoE) is a popular technique to improve capacity of Large Language Models (LLMs) with conditionally-activated parallel experts. However, serving MoE models on memory-constrained devices is challenging due to the large parameter size. Typical solutions such as memory swapping or expert pruning may lead to significantly higher latency or severe accuracy loss.
In this paper, we introduce SwapMoE, a framework for efficient serving of MoE-based large language models with tunable memory budgets. The main idea of SwapMoE is to keep a small dynamic set of important experts, namely Virtual Experts, in the main memory for inference, while seamlessly maintaining how the Virtual Experts map to the actual experts. Experiments have shown that SwapMoE can reduce the memory footprint while maintaining reasonable accuracy. For example, on text summarization tasks with Switch Transformer, SwapMoE can reduce the memory consumption from 14.2 GiB to 4.7 GiB, together with 50\% latency reduction and a slight Rouge-2 score drop of 0.041.

\end{abstract}

\section{Introduction}

Recently, the world has witnessed the great advancement of pre-trained large language models~\cite{li2024personal}.
Such an advancement is driven by the phenomenon that the larger model capacity generally leads to higher intelligence \cite{scalinglaw:arxiv20:Kaplan}.
Among various attempts to scale up neural networks, Mixture of Experts (MoE) \cite{vanilla_moe:iclr17:Noam} is a promising technique that can avoid linearly increasing computation based on conditional sparse computation.
However,  memory-costrainted devices is often a major concern in edge AI training and serving \cite{huang2023elastictrainer,10.1145/3498361.3539765,10.1145/3498361.3538928,ma2023costeffective,rui2023convrelu}. The constraint is even more challenging for large MoE models. 
For example, a Switch Transformer~\cite{switch_moe:jmlr22:Fedus} with 64 experts per layer requires 14 GiB of memory for inference, which is impossible to fit in consumer devices that typically have only 8 GiB, 4 GiB, or lower memory size. What's more, many consumer devices need to serve multiple applications, and the system-allocated memory budget for each application is even more limited.


Researchers have proposed various methods to reduce the memory footprint of MoE model inference.
The most straightforward approach is to use memory swapping, \ie dynamically loading/unloading the parameters from/to the external memory.
For example, \citet{cachemoe:arxiv23:huang} propose to load experts into the memory on demand to reduce memory footprint but introduce additional latency overhead, 
EdgeMoE~\cite{yi2023edgemoe} focuses on memory swapping in the decoding phase of MoE-based language models and reduces the swapping overhead using quantization. However, it only considers the scenario of single token inference, which does not align with real-world use cases.
Other algorithm-perspective approaches~\cite{taskexpertpruning:arxiv22:Chen,kim2021scalable} propose to prune the experts to reduce model size.
However, swapping-based approaches either have to trade latency for a reduced memory footprint, while pruning-based methods would lead to accuracy loss and require model training.
How to efficiently serve an off-the-shelf MoE model under memory constraints remains challenging.


Fortunately, there is a unique property in many MoE workloads that can be exploited for performance optimization - activation locality.
For example, in generative language models, the output tokens are produced one by one, and they mostly belong to the same semantic context. The successive queries of a user are also semantically related.
Such locality can potentially lead to consistent activation patterns of MoE experts, therefore create room for more intelligent expert management.

\textbf{Our work.} We introduce \name to enable efficient continuous MoE serving under memory constraints.
The key idea is to maintain a dynamically-updated compact set of \core in the main memory for MoE inference, instead of the redundant set of all experts in the original model. The weights of \core are seamlessly updated according to the data distribution locality and profiled hardware capabilities. As such, the memory footprint and latency of each individual inference process in \name are the same as running a smaller MoE model, while the advantage of large-capability MoE remains since each expert still has the chance to participate in the computation.

We implement \name with Huggingface Transformers~\cite{wolf2019huggingface} library.
To evaluate the effectiveness of \name, we conduct experiments with Switch Transformer~\cite{switch_moe:jmlr22:Fedus} (SwitchT for short), and GPTSAN~\cite{gptsan} models on natural language processing tasks, compared with the normal model inference scheme and strong baselines.
For example, on text summarization tasks with Switch Transformer, \name can reduce the memory consumption from 14.2 GiB to 4.7 GiB (67\% less), together with 50\% latency reduction and a slight Rouge-2 score drop of 0.041, which shows that \name enables the serving of large MoE models on resource-constrained consumer devices with limited memory budgets.

We summarize our key technical contributions as follows:
\begin{itemize}[left=0.5em]
\item We propose a novel MoE inference framework that supports efficient serving of large MoE models under memory constraints. Our method enables deployment of off-the-shelf MoE models to consumer devices with tunable memory budgets.
\item We investigate several properties of MoE models, including the expert activation locality, per-sample expert importance, and layer-wise tolerance to absent experts, which can potentially benefit other works on MoE optimization.
\item We implement \name with popular inference frameworks and conduct extensive experiments with large MoE models on consumer devices. The results have demonstrated the effectiveness of our approach on natural language processing tasks.
\end{itemize}
\section{Background and Motivation}
\label{section:background}

The sparse MoE~\cite{vanilla_moe:iclr17:Noam} is the most commonly used, in which only one or few (represented as $k$) experts are activated for each input. Specifically, in a sparse MoE with $k=1$, only the $i$-th expert with max $G(x)_i$ is activated for input $x$, and the output $y=G(x)_i E_i(x)$.

The MoE structure is often accompanied by the Transformer architecture \cite{m3vitmoe:acl22:liang,moefication:acl22:zhengyan}, in which the input of each MoE layer is a sequence of tokens and each token may choose different experts in one MoE layer.

\subsection{Limitations of Conventional Solutions}

\begin{figure}[]
    \centering
        \includegraphics[width=0.35\textwidth]{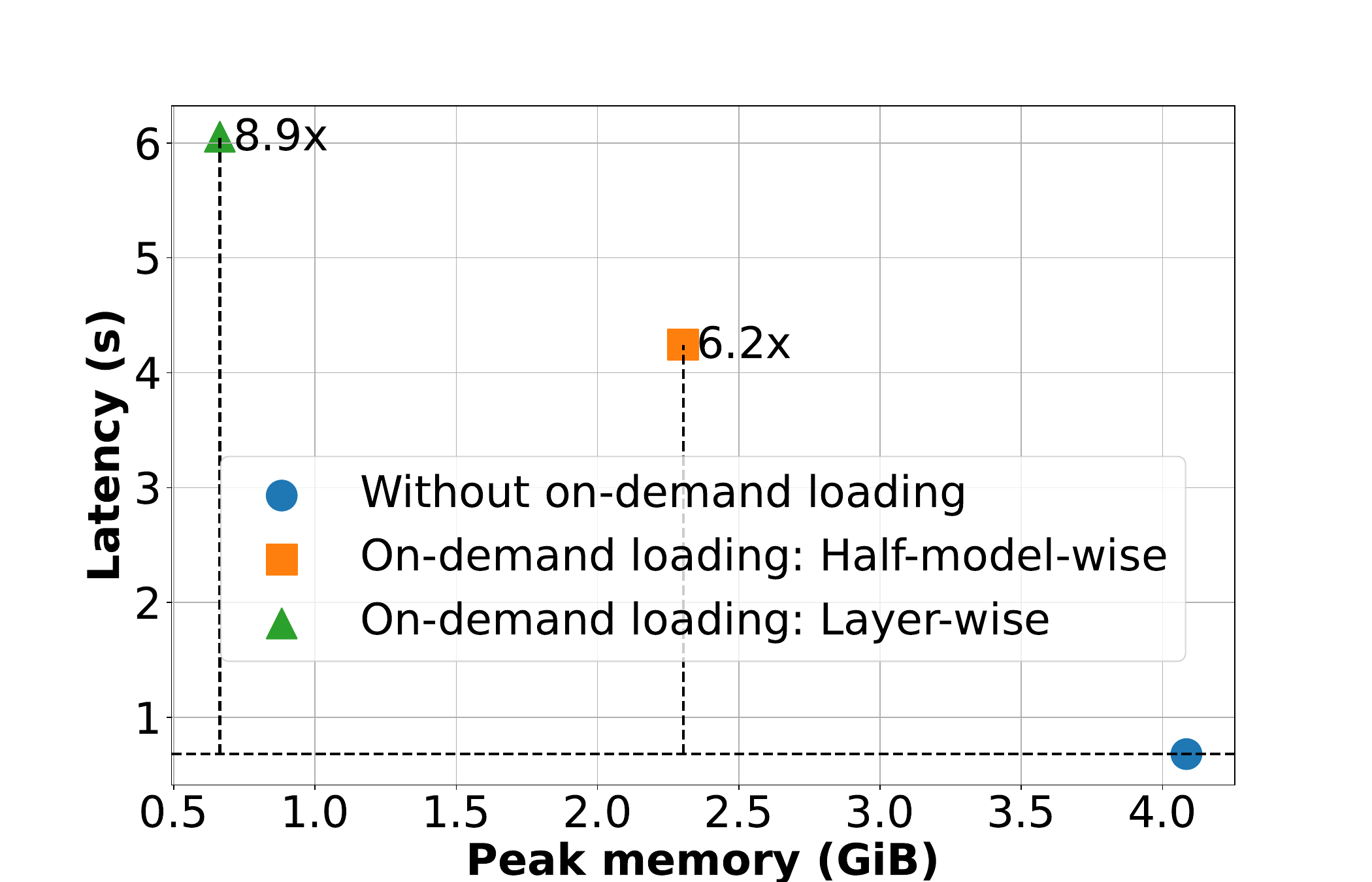}
        \vspace{-5pt}
        \caption{SwitchT-16: Naive on-demand expert loading reduces memory, but also results in huge inference overhead.}
    \label{figure:layer_wise_loading_latency_vs_memory}
\end{figure}

\begin{figure}[]
    \centering
        \includegraphics[width=0.35\textwidth]{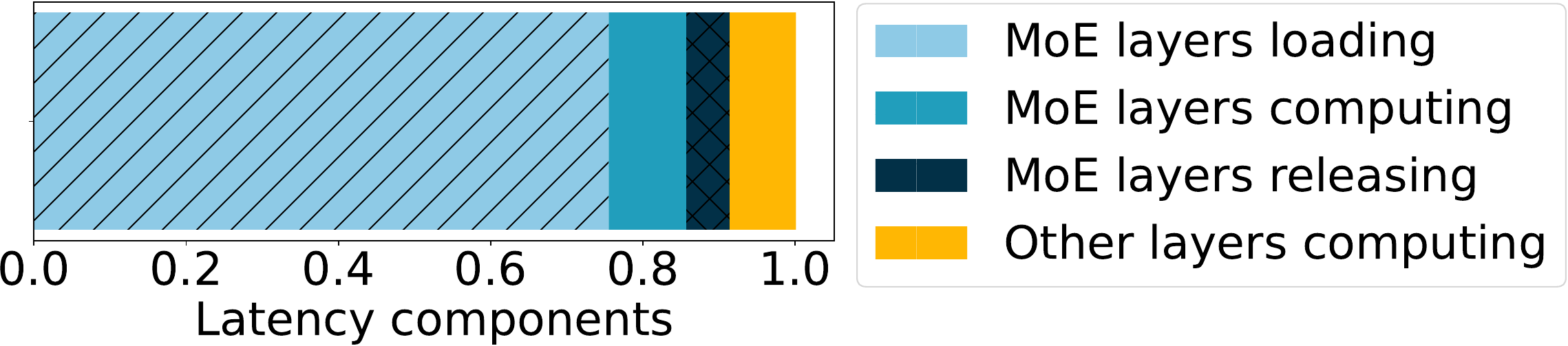}
        \vspace{-5pt}
        \caption{SwitchT-16: Latency breakdown of MoE model inference with layer-wise memory swapping. The transmission of model weights consumes the majority of the time.}
    \label{figure:layer_wise_loading_latency_breakdown}
\end{figure}

\begin{figure}[]
    \centering
        \includegraphics[width=0.45\textwidth]{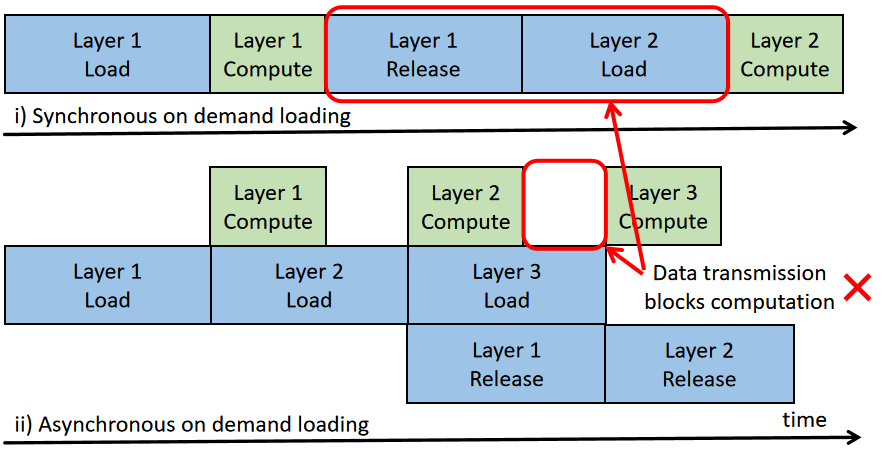}
        \vspace{-10pt}
        \caption{
        Weight loading may block computation when running MoE with layer-wise memory swapping. Due to the large size of MoE layers and sparse computation, loading the weights of a layer is always slower than computing the layer, which slows down the inference even if the weights are loaded asynchronously.
        }
    \label{figure:offloading_process}
\end{figure}

\begin{figure*}[]
    \centering
        \includegraphics[width=0.85\textwidth]{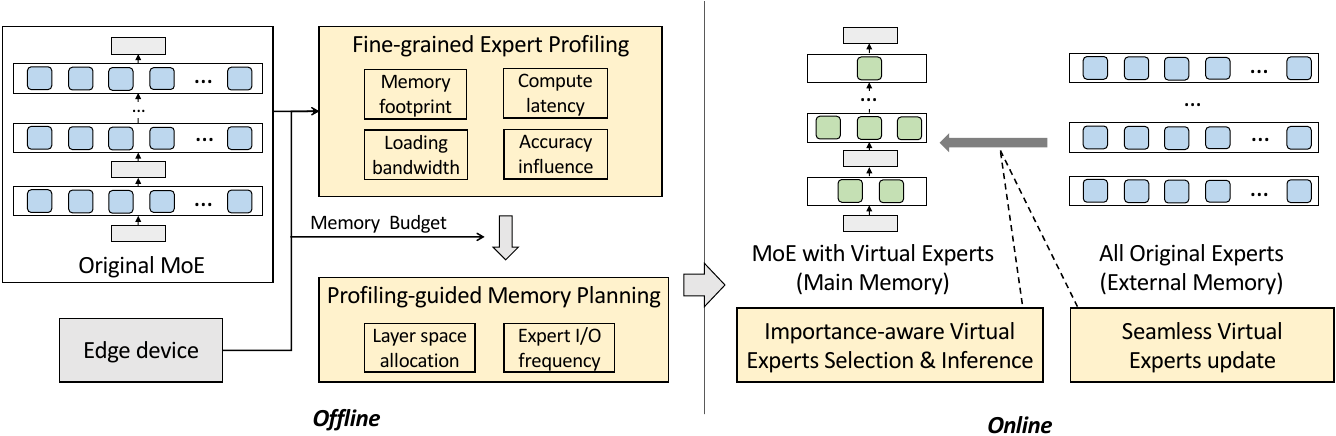}
        \caption{The workflow of \name.
        Given an off-the-shelf MoE model and a memory-constrained consumer device, we satisfy the constraint by executing the model with a smaller set of experts (Virtual Experts). The Virtual Experts are selected, used and updated seamlessly at runtime, and the memory allocation for the experts is determined at offline.
        }
        \vspace{-10pt}
    \label{figure:algo overview}
\end{figure*}

Existing methods for reducing the resource overhead of MoE model inference include on-demand loading \cite{lane2016deepx} (\ie loading the parameters of MoE layers when they are needed and release afterwards) and MoE model pruning \cite{taskexpertpruning:arxiv22:Chen,moedistillation:arxiv21:Kudugunta} (\ie cutting the less important experts permanently). 

While the on demand loading method can reduce the GPU memory usage during MoE model inference without affecting model accuracy, it introduces significant latency overhead with each MoE layer parameter transmission via PCIe. 
In Figure~\ref{figure:layer_wise_loading_latency_vs_memory}, running MoE models with on-demand loading introduces 6.2x-8.9x higher latency.
Meanwhile, the transmission of a large amount of parameter data can significantly impede the computation process.
As shown in Figure~\ref{figure:layer_wise_loading_latency_breakdown} and Figure~\ref{figure:offloading_process}, model parameters transmission takes up most inference time and significantly obstructs the model's computational process and may deplete IO resources. 

The pruning-based method directly reduces the model's parameter and computational load, thereby lowering the GPU memory usage and inference latency during MoE model inference. However, it greatly compromises the model's performance due to compromised the models' capacity.
For example, when utilizing expert pruning to reduce memory usage by 30\% on SwitchT-32, the model's accuracy decreased by 14\%.
\name combines the advantages of both methods - we try to reduce the latency overhead of memory saving while striving to maintain the model accuracy at the same time.

\subsection{Activation Locality in MoE Models} \label{sec:goal_challenges}

Data distribution locality is an important characteristic in many AI applications, which refers to the phenomenon that successive input samples are distributionally similar or correlated. 
Firstly, each inference process of the model produces a token, and the successive tokens belong to the same sentence.
Secondly, AI models are usually deployed in a fixed environment and used for serving an individual user or organization, the successive input samples (\eg conversations) are semantically close to each other.

Since each expert in a MoE model is trained to handle certain data distribution, there exists an opportunity to cache the most relevant experts in the main memory at each time step, therefore reducing the memory consumption. The data distribution locality in AL applications further produces the change to reuse the cached experts for successive input samples, which can reduce the overhead of parameter loading.
\section{Our Design: \name} \label{sec:approach}

\begin{figure}[]
    \centering
        \includegraphics[width=0.5\textwidth]{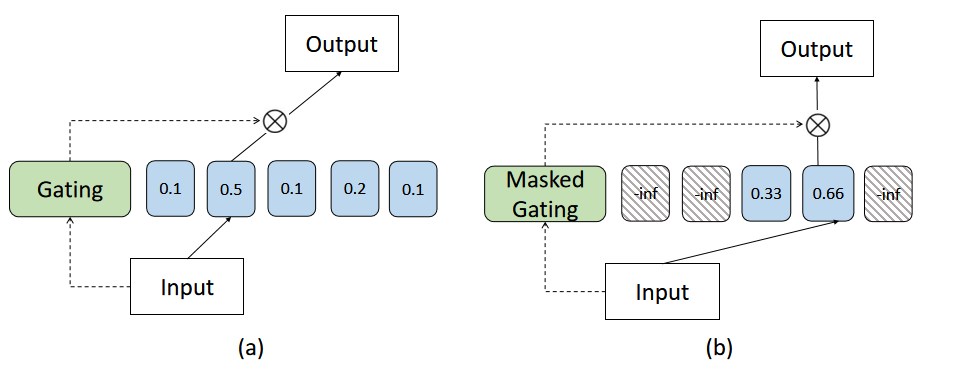}
        \vspace{-10pt}
        \caption{
            (a) Original gating of MoE: all experts may by used for inference;
            (b) Ours Masked Gating: only \core will be used.
        }
        \vspace{-10pt}
    \label{figure:maksed_gating}
\end{figure}

\begin{figure}[]
    \centering
        \includegraphics[width=0.5\textwidth]{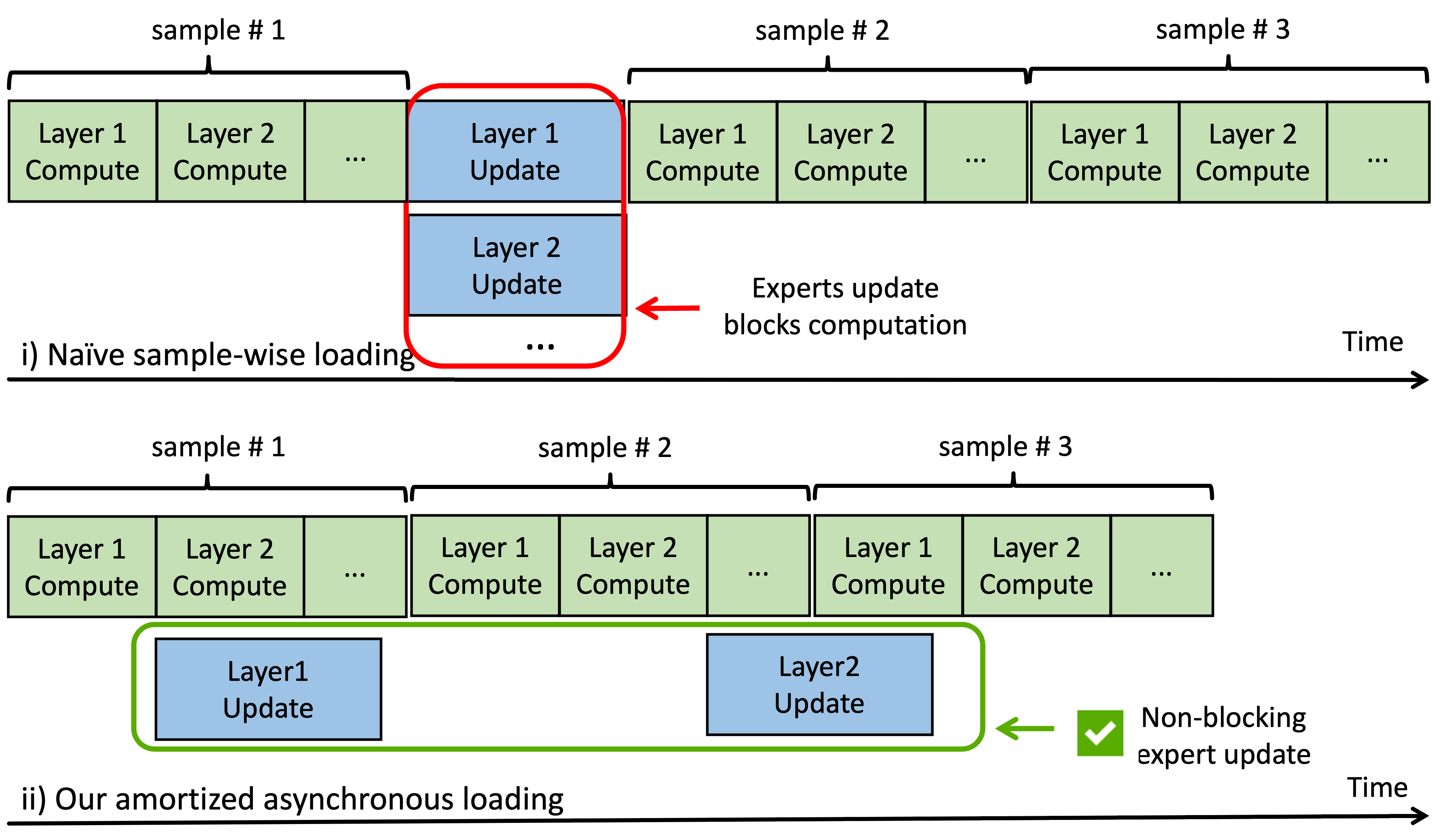}
        \caption{
            Different loading strategy of \core. i) synchronously updating all experts after single sample inference; ii) amortized asynchronous expert loading across different samples.
        }
        \vspace{-10pt}
    \label{figure:amortized_update}
\end{figure}

\name utilizes a two-phase holistic design, as shown in Figure~\ref{figure:algo overview}. 
In the online phase, the job of \name is to efficiently identify, update, and use a subset of experts (\core) for memory-constrained MoE inference.
The job of the offline phase is to obtain an optimal memory plan for the online phase to facilitate efficient and accurate MoE inference. 

\subsection{Importantance-aware \core Selection \& Inference}\label{sec:moe_inference}
Firstly, we must wisely select the most important experts and update them according to the data distribution to maintain accuracy with \core-based MoE inference.
To address this problem, we introduce the concept of the \textbf{expert importance score},
which can be written as
\begin{equation}
    importance(E_i,X) = \sum_{x}^{\mathcal{X}_i}  ||x||*||G(x)_{i}||*||E_{i}||, 
\label{equation:importance}
\end{equation}
where $\mathcal{X}_i$ is the set of tokens passed to expert $E_i$ during the inference process of $X$. This score only involves simple magnitude computation thus can be calculated efficiently.


As shown in Figure~\ref{figure:maksed_gating}, we use Masked Gating to redirect all inference requests to \core.

\subsection{Seamless \core Update}\label{sec:expert_loading}
Once we have the importance scores for all experts, we can update \core accordingly at runtime by loading the important experts into the main memory and the unimportant experts out.
We introduce two techniques to reduce overhead, including \textbf{amortized expert loading} and \textbf{asynchronous expert loading}, as shown in Figure~\ref{figure:amortized_update} (ii).

\subsection{Fine-grained Expert Profiling}\label{sec:expert_profiling}

To facilitate efficient and accurate model inference, it is necessary to know the performance of \core given a specific hardware and configuration and determine what kind of configuration will lead to better perfromance. 
We conduct fine-grained expert profiling in advance, gather information related to hardware memory usage, inference latency, accuracy, and IO bandwidth, and establish the relationship between Virtual Experts configurations and performance, 
Specifically, configurations are:
\begin{align}\label{equation:configuration}
    \text{config} & = \{frequency, \text{num\_experts}\} \nonumber \\
    \text{num\_experts} & = \{\#experts_1, ..., \#experts_L\},
\end{align}
where $frequency$ represents the update frequency of the experts in the core, indicated as the number of inputs between each pair of updates. $\#experts_i$ denotes the number of experts to be retained in the core for the $i$-th layer. 

\textbf{Problem Formulation.}
The primary objective of offline planning is to identify the optimal configuration $\hat{\text{config}}$ that meets memory constraints while maximizing the model's accuracy and minimizing inference latency. 
Formally, the process can be described as follows:
\begin{align}\label{equation:formulation}
maximize\ & {E_\text{accuracy}(\text{config})}, \nonumber \\
minimize\ & {E_\text{latency}(\text{config})}, \nonumber \\
    \text{s.t.} \quad &E_\text{memory}(\text{config})\leq {LIMIT}_\text{memory}, 
\end{align}
where $E_\text{accuracy}$, $E_\text{memory}$, and $E_\text{latency}$ are used to estimate the accuracy, memory footprint, and inference latency of a model under a specific configuration. ${LIMIT}_\text{memory}$ represent the constraints on main memory. 


\subsection{Profiling-guided Memory Planning}\label{sec:expert_planning}
Section~\ref{sec:moe_inference} and Section~\ref{sec:expert_loading} have shown how the \core are selected and used for inference at runtime. However, many questions remain unanswered. For example, how to distribute \core across different layers, how to allocate limited memory size to different layers, and how to make the best use of limited memory bandwidth and enabling frequent updates of experts without blocking computations, etc. These questions are crucial to satisfy the memory constraint and minimize accuracy loss. They are addressed through a profiling-guided memory planning, which using profiling information obtained from Section~\ref{sec:expert_profiling}.

\subsubsection{Expert I/O Frequency}
The update frequency affects the usage of IO bandwidth, and if too much data need to be transferred through IO, it will cost a long time and block the computation. 
In other words, when the experts in \core are no longer important for the current sample, we need to promptly replace them with the latest important experts. A higher update frequency is naturally preferable, but increased frequency can escalate hardware IO resource consumption, potentially even obstructing inference computation and leading to increased latency. At the same time, because a higher frequency is more beneficial for dynamically maintaining the most important experts, it is important to increase the frequency as much as possible without blocking the computation.
Our strategy involves testing from a low update frequency to a high update frequency until we identify the inflection point at which the update frequency affects the inference latency, allowing us to select the optimal update frequency.

\subsubsection{Layer Space Allocation}
To allocate the limited memory budget to different layers, allow layers with more memory to utilize more Virtual Experts, find the optimal configuration ($\hat{\text{config}}$) that maximizes the model accuracy while satisfying given memory constraints, we utilize memory planner to obtain the optimal memory allocation scheme based on the previously obtained performance model obtained from $E_\text{accuracy}$, $E_\text{memory}$, and $E_\text{latency}$ in Section~\ref{sec:expert_profiling}.

A naive method to find the optimal configuration is to iterate over all possible configurations and keep track of the best-performing one.
However, we found that this approach does not fully leverage the modeled functions to find the optimal configuration due to the enormous search space. For instance, the search space in a 12-layer SwitchT-16 would be $12^{16}$. 

Consequently, we employ the genetic algorithm~\cite{holland1992genetic} for the search process.
Specifically, we initialize a set of configurations randomly, and iteratively update them based on their performance metrics estimated with $E_\text{accuracy}$, $E_\text{memory}$ and $E_\text{latency}$. In each iteration, we randomly change one parameter in each configuration and create new configurations by exchanging or averaging existing configurations. The configurations that violate resource constraints or yield suboptimal performance are removed.
We observed that certain layers in the model may have a more significant impact on model performance. The genetic algorithm can perceive this characteristic and preserve experts that have a greater influence on performance. 
In Section~\ref{sec:experiment}, we will illustrate some configurations found by the algorithm.

\section{Evaluation}\label{sec:experiment}


\subsection{Experimental Setup}
\textbf{Platforms.}
We use two devices: a Jetson Nano and a Jetson AGX ORIN.
The batch sizes are all set to 1 as in common edge-side continuous serving scenarios.
We use different latency and memory budgets to simulate different resource constraints.

\textbf{Tasks, datasets and models.}
We evaluate the performance of \name on two common DL tasks:

\textbf{Summarization} aims to summarize long texts into short texts. We select the most popular summarization model, Switch Transformer~\cite{switch_moe:jmlr22:Fedus} (16, 32, and 64 experts per layer, denoted as SwitchT-16, SwitchT-32 and SwitchT-64) and use the samsum dataset~\cite{imagenetvid:ijcv2015:Russakovsky} and report Rogue-2 accuracy, where higher Rogue-2 accuracy means better.

\textbf{Language modeling} aims to predict the next word given the previous words. We use GPTSAN~\cite{gptmoe:acl22:Artetxe} as the base model and use the Wikipedia-japanese~\cite{wikipedia-japanese} dataset for evaluation. The performance of the language model is measured by perplexity, where lower perplexity means better.

The pre-trained weights of the Switch Transformers and GPTSAN are obtained from the official Huggingface Transformer repository. All datasets can be downloaded from public websites.

\textbf{Baselines.}
The basic baseline is original MoE model.
To show the superiority of \name, we also compare it with the following baselines:
`\textbf{Pruning}' (keeping a certain portion of experts with the largest magnitude $||E_i||$ in each layer, no switching in/out), `\textbf{On demand}'~\cite{cachemoe:arxiv23:huang} (keep a certain portion of experts in main memory, load the requested expert from memory on demand).
`\textbf{Pruning}' is akin to a simplified version of pruning-based methods without training, while `\textbf{On demand}' is a simplified version of swapping-based approaches.

\begin{figure*}
    \centering
    \includegraphics[width=0.6\textwidth]{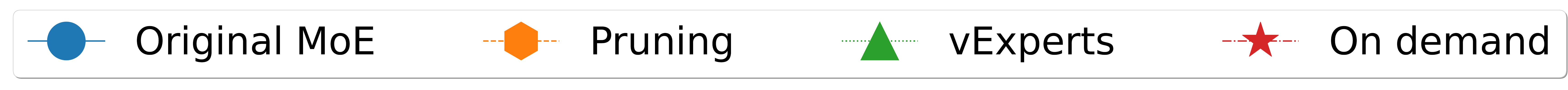}
    \vspace{-10pt}
\end{figure*}

\begin{figure*}
    \begin{subfigure}[]{0.2\textwidth}\centering
        \includegraphics[width=0.99\textwidth]{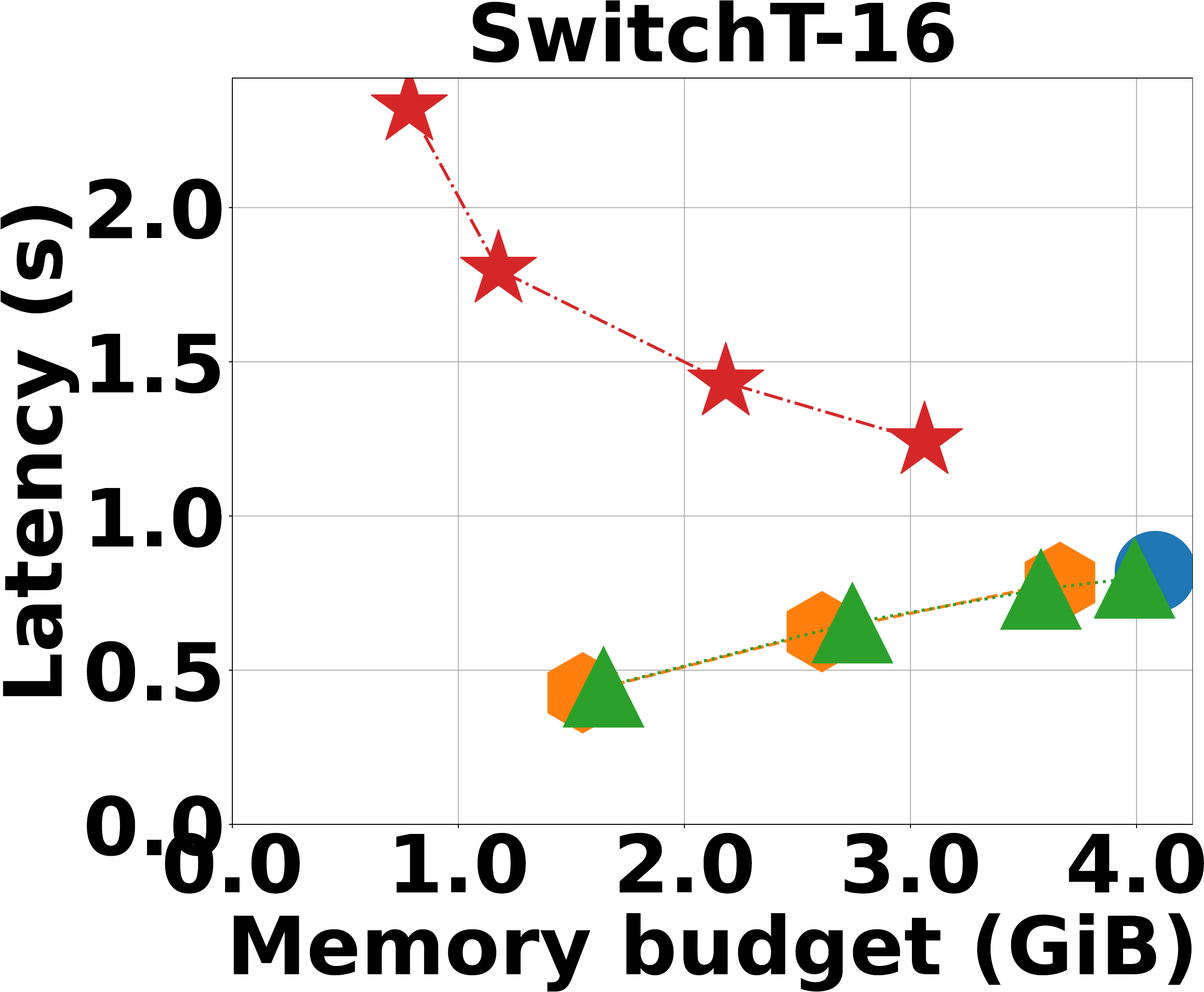}
    \end{subfigure}
    \hfill
    \begin{subfigure}[]{0.2\textwidth}\centering
        \includegraphics[width=0.99\textwidth]{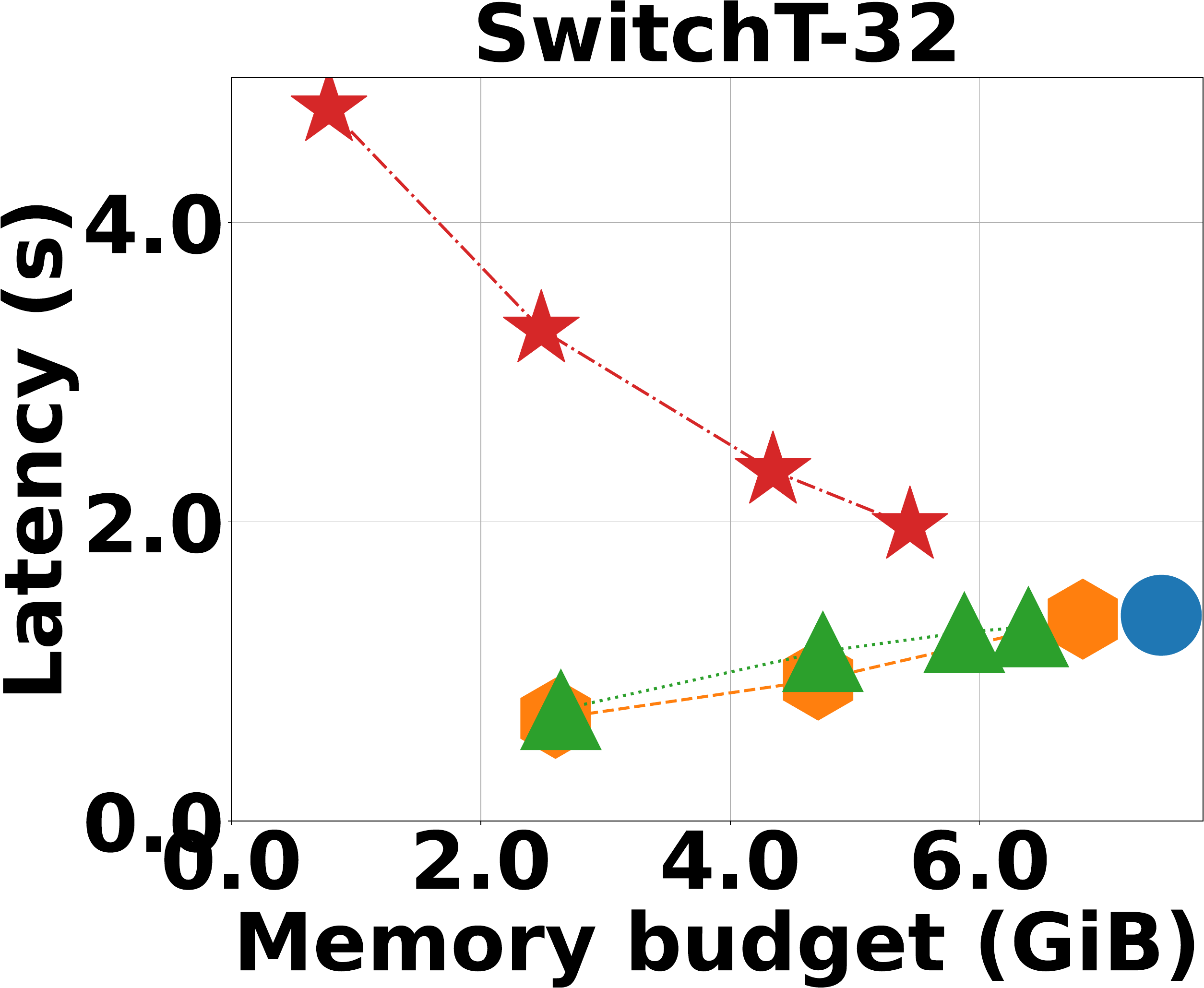}
    \end{subfigure}
    \hfill
    \begin{subfigure}[]{0.2\textwidth}\centering
        \includegraphics[width=0.99\textwidth]{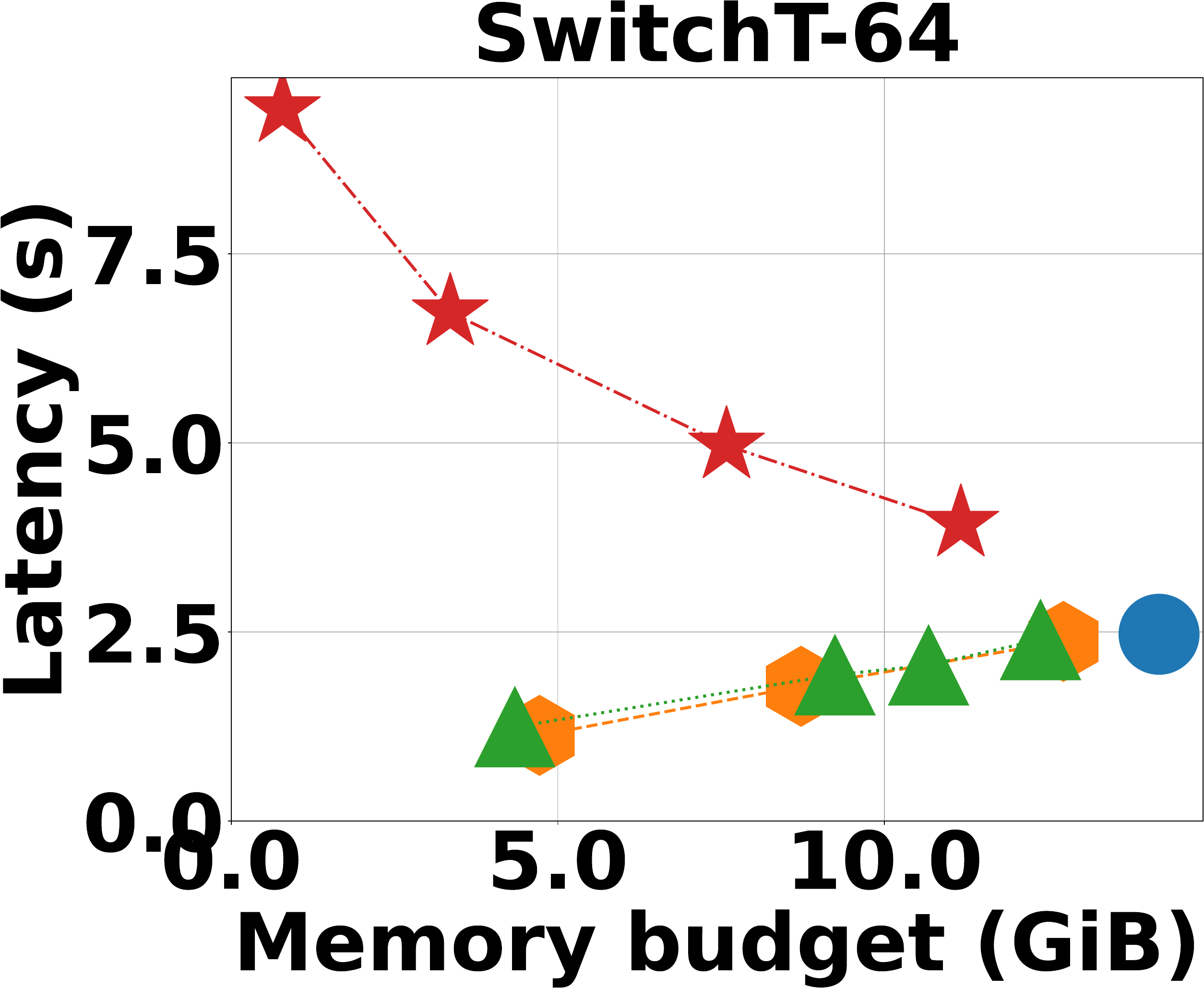}
    \end{subfigure}
    \hfill
    \begin{subfigure}[]{0.2\textwidth}\centering
        \includegraphics[width=0.99\textwidth]{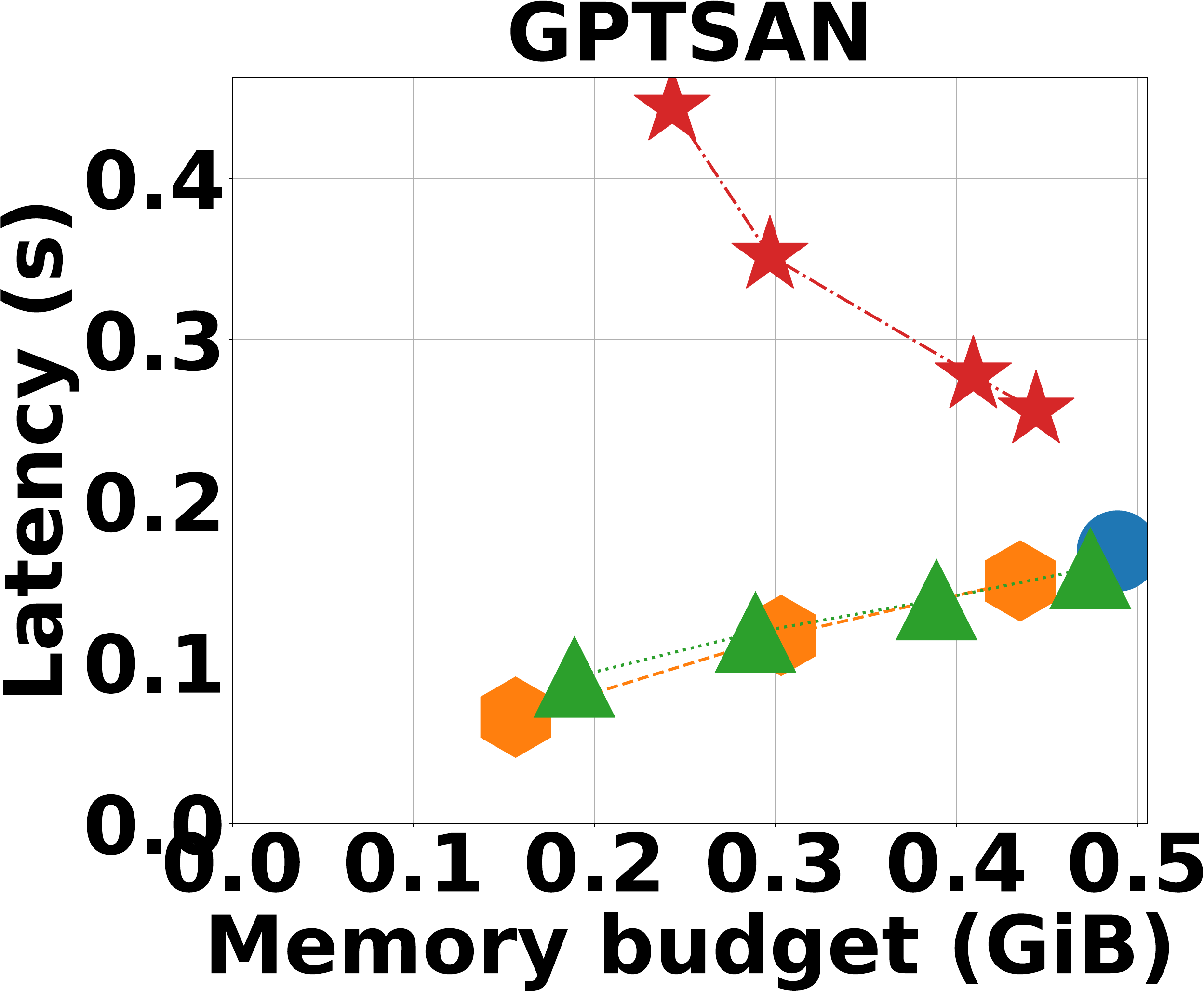}
    \end{subfigure}
    \caption{
        End-to-end latency achieved by \name and baselines under different memory budgets on Jetson AGX.
        \name achieves similar latency with the pruned compact models while with much more parameters in use.
    }\label{result:memory-latency-tradeoff}
    \vspace{-15pt}
\end{figure*}


\begin{figure*}
    \begin{subfigure}[]{\textwidth}\centering
        \begin{subfigure}[]{0.2\textwidth}\centering
            \includegraphics[width=0.99\textwidth]{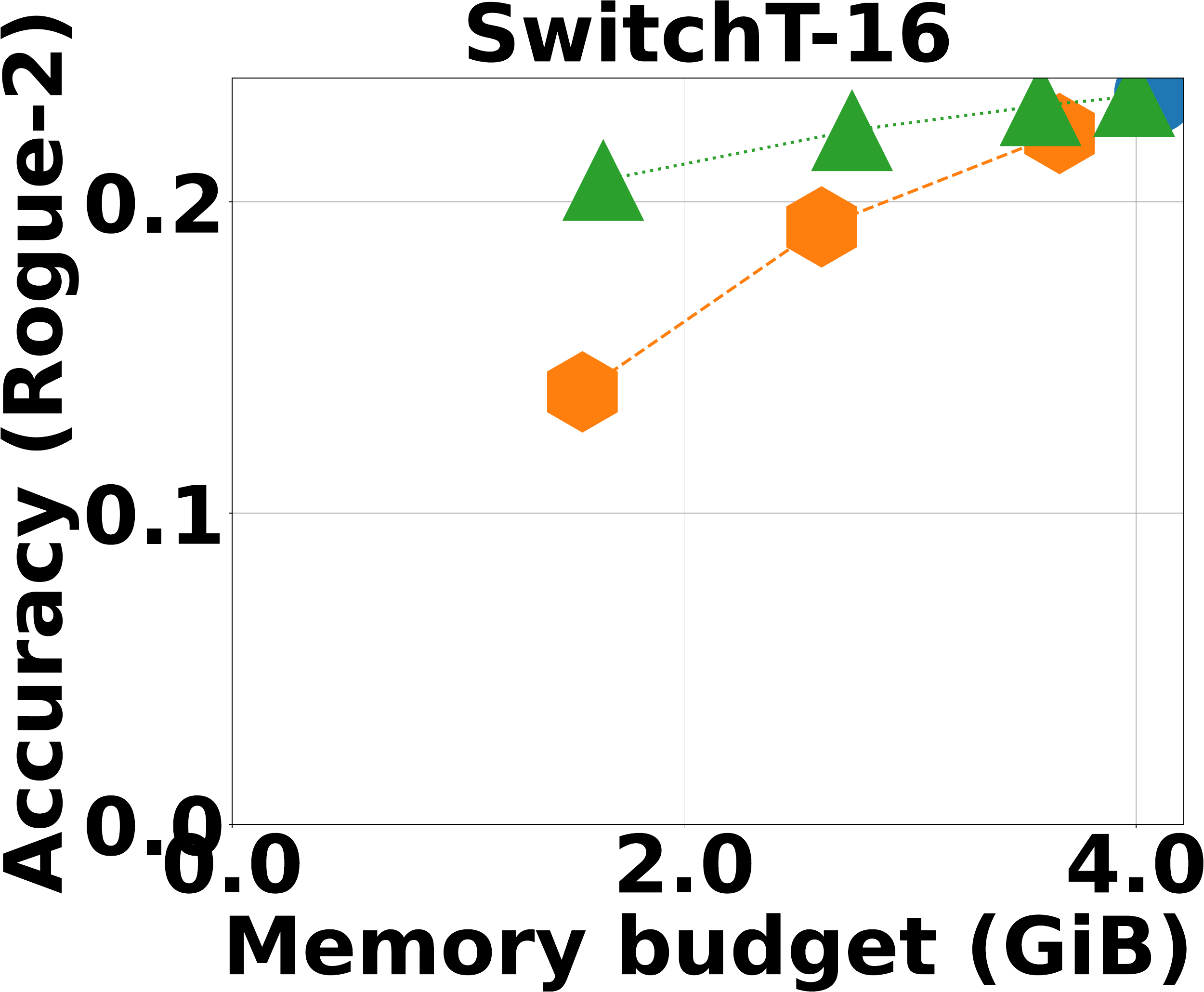}
        \end{subfigure}
        \hfill
        \begin{subfigure}[]{0.2\textwidth}\centering
            \includegraphics[width=0.99\textwidth]{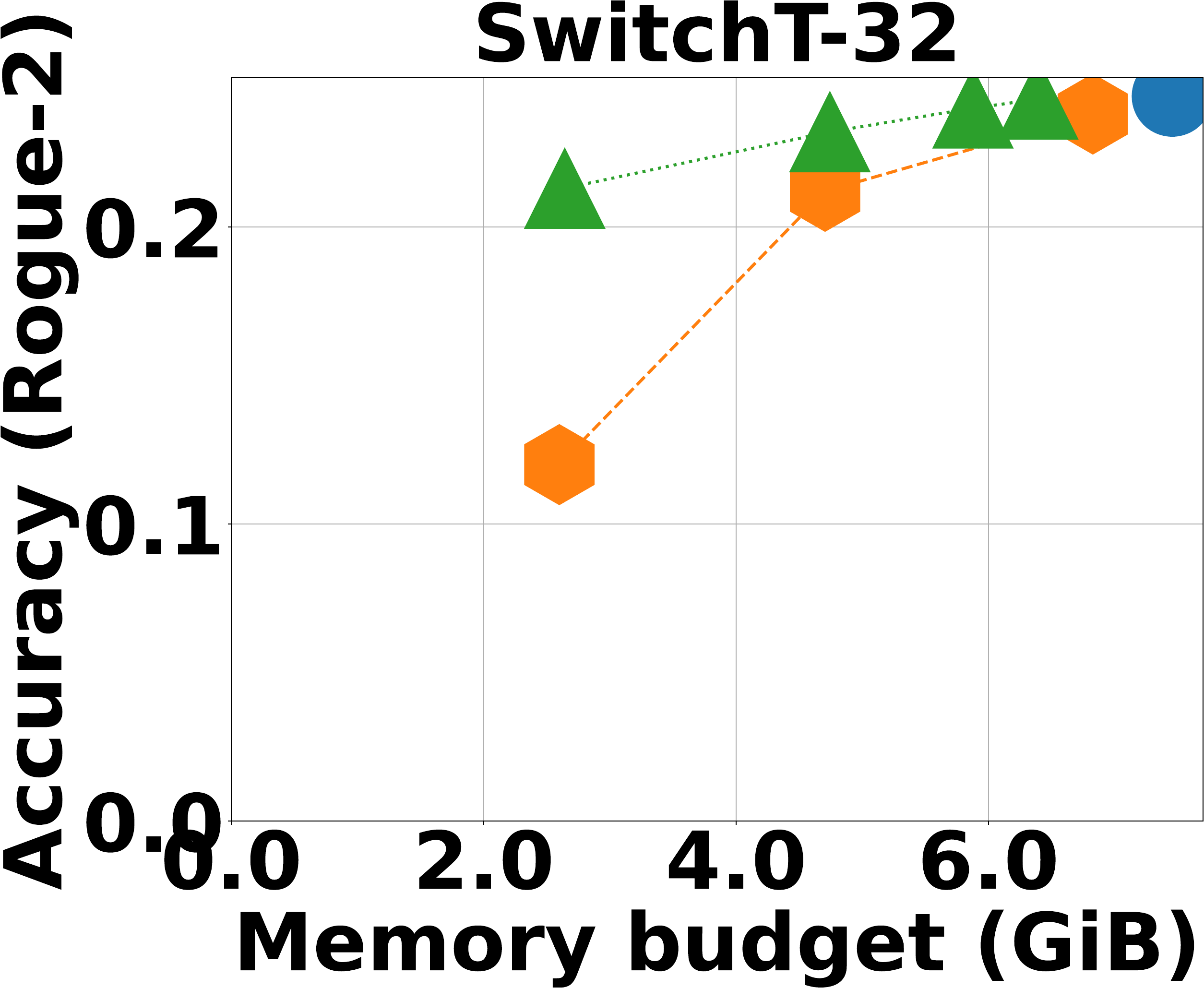}
        \end{subfigure}
        \hfill
        \begin{subfigure}[]{0.2\textwidth}\centering
            \includegraphics[width=0.99\textwidth]{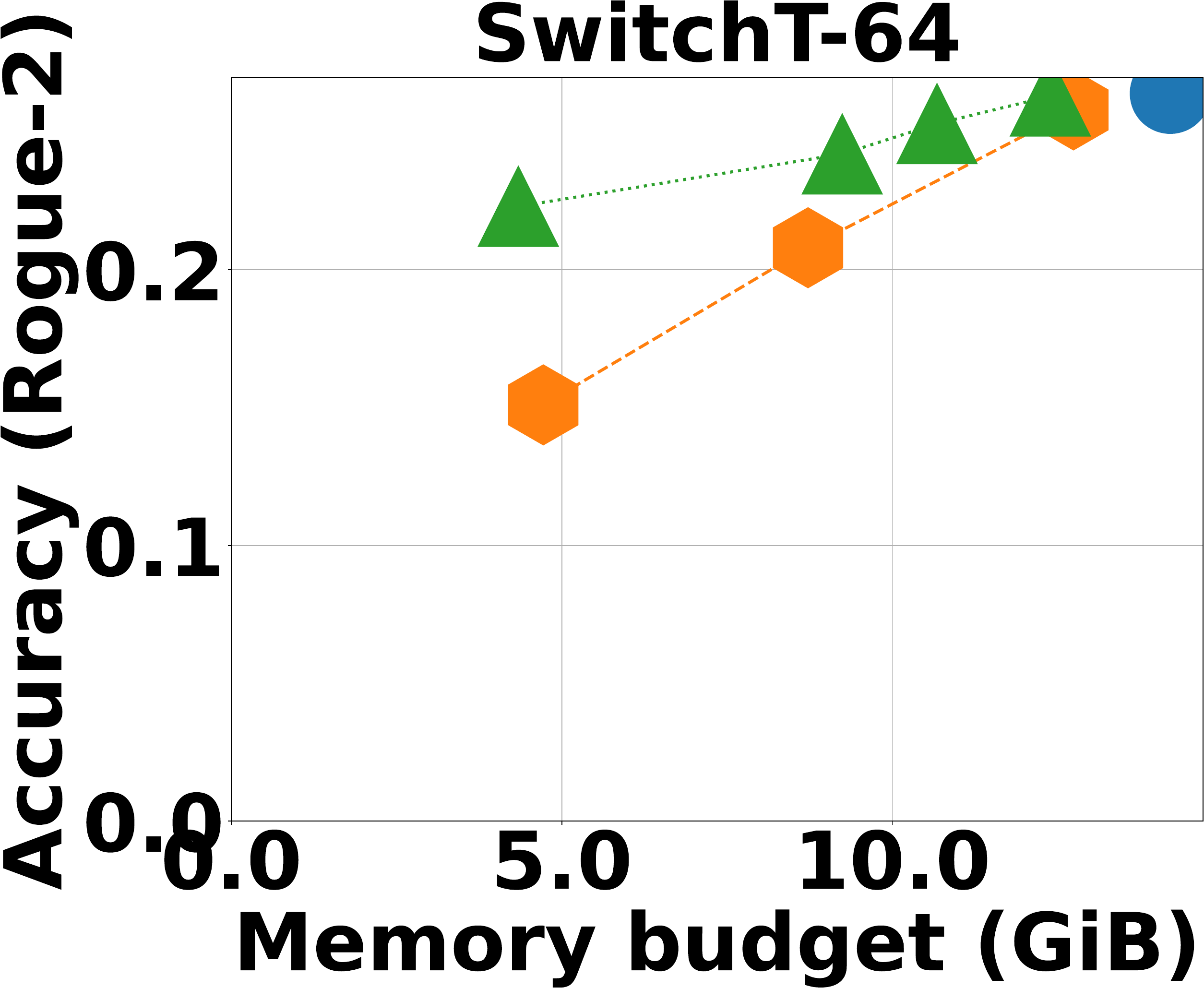}
        \end{subfigure}
        \hfill
        \begin{subfigure}[]{0.2\textwidth}\centering
            \includegraphics[width=0.99\textwidth]{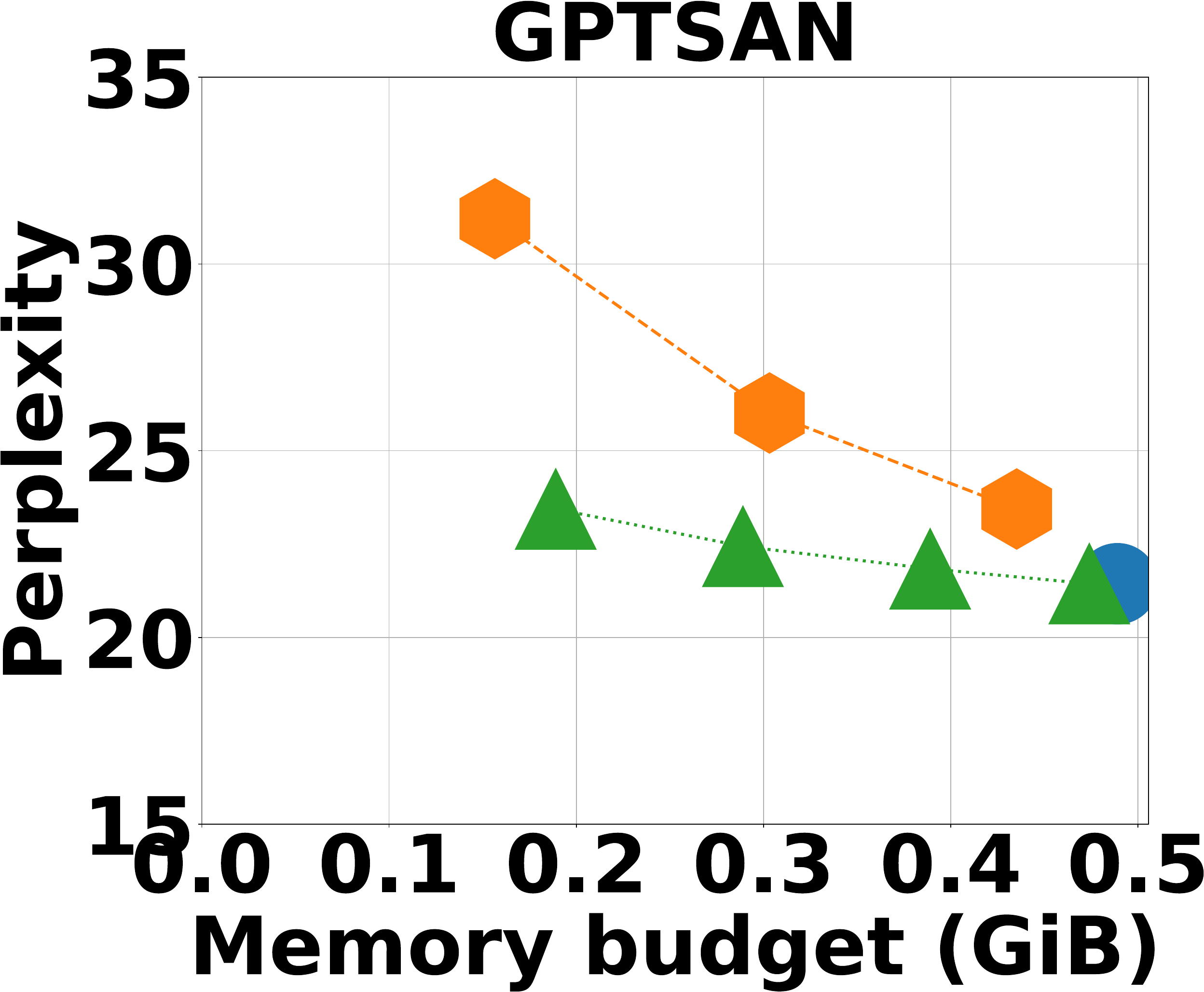}
        \end{subfigure}
        \caption{
         Jetson AGX
        }\label{result:memory-accuracy-tradeoff:agx}
    \end{subfigure}
    \begin{subfigure}[]{\textwidth}\centering
        \begin{subfigure}[]{0.2\textwidth}\centering
            \includegraphics[width=0.99\textwidth]{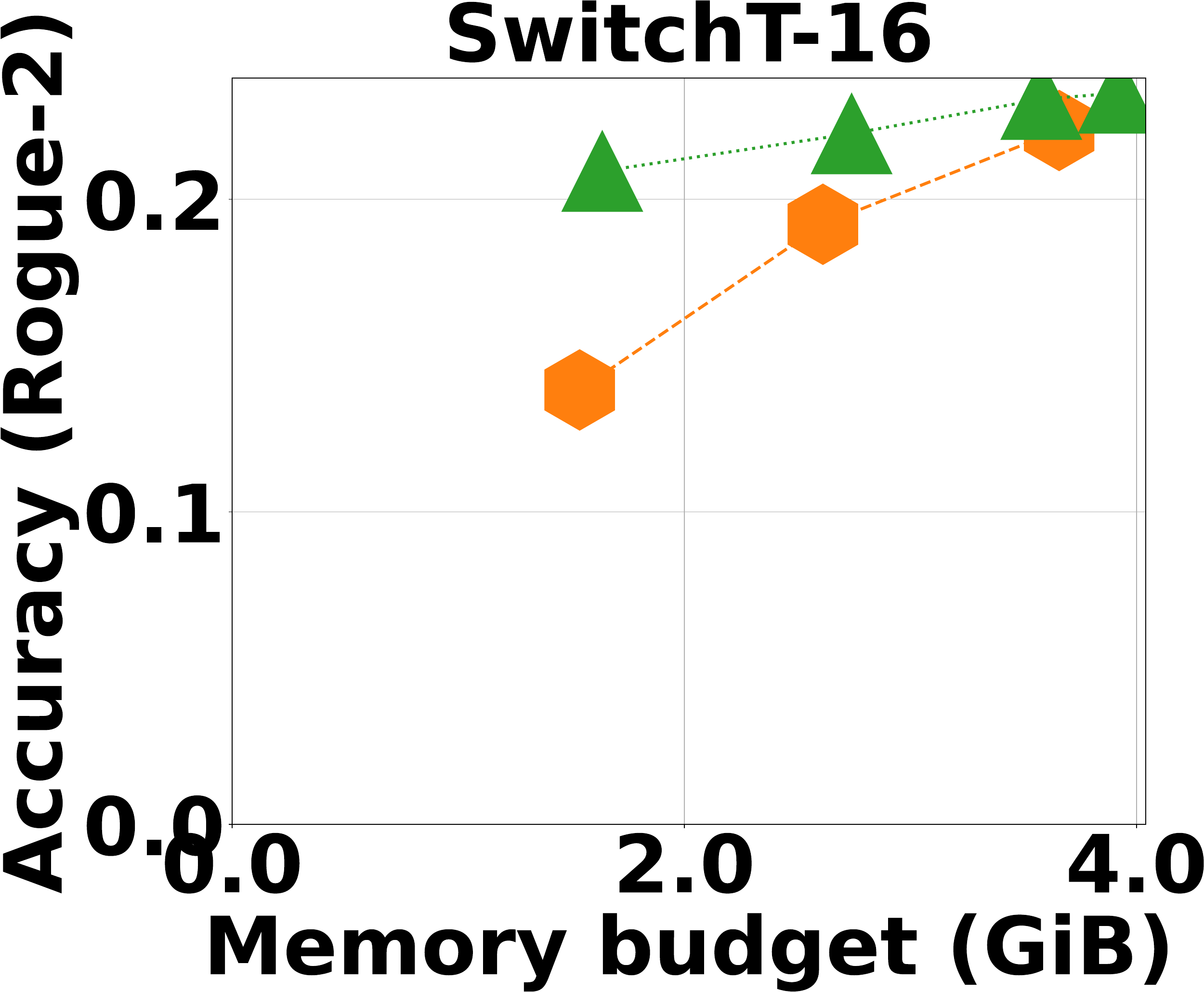}
        \end{subfigure}
        \hfill
        \begin{subfigure}[]{0.2\textwidth}\centering
            \includegraphics[width=0.99\textwidth]{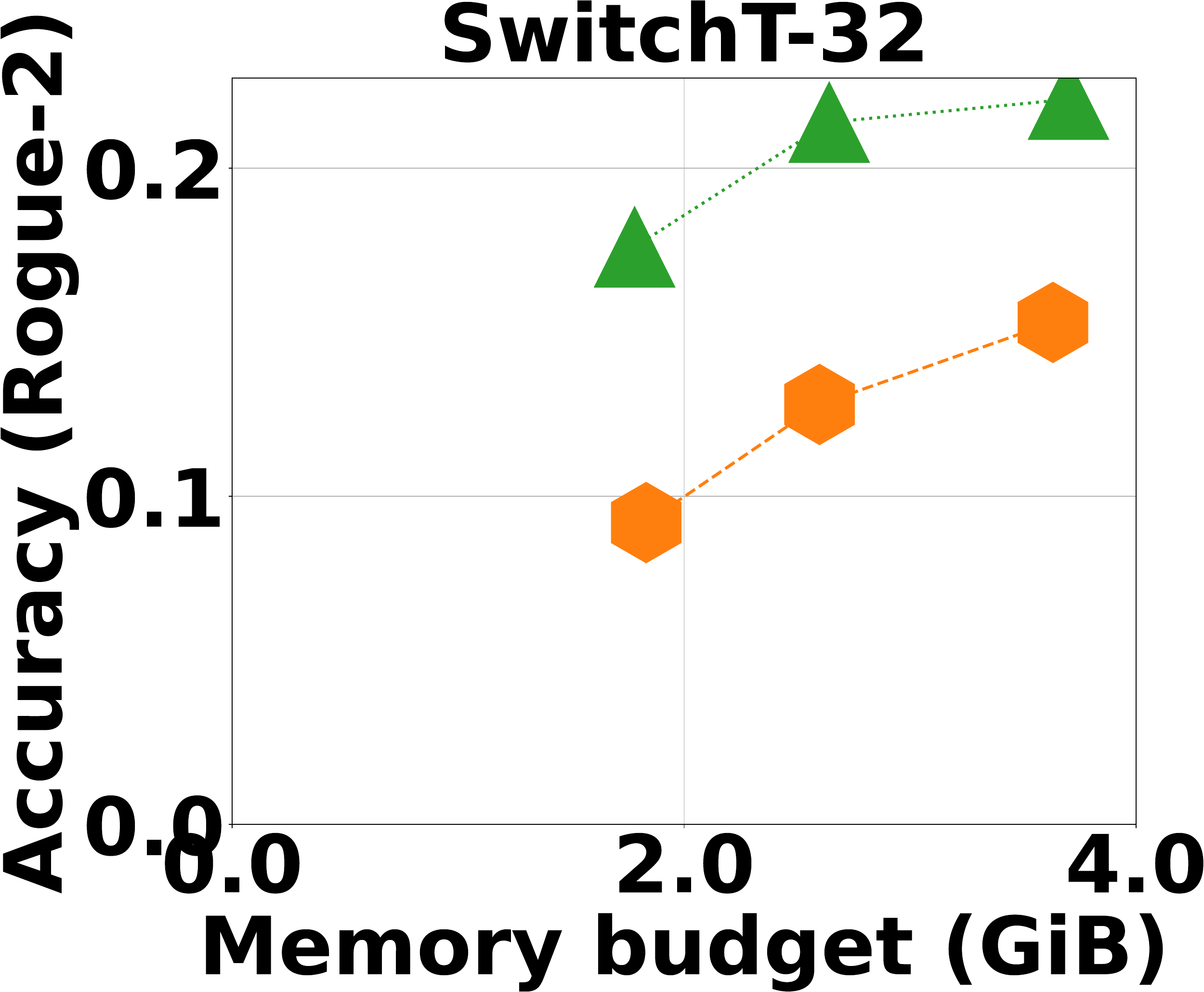}
        \end{subfigure}
        \hfill
        \begin{subfigure}[]{0.2\textwidth}\centering
            \includegraphics[width=0.99\textwidth]{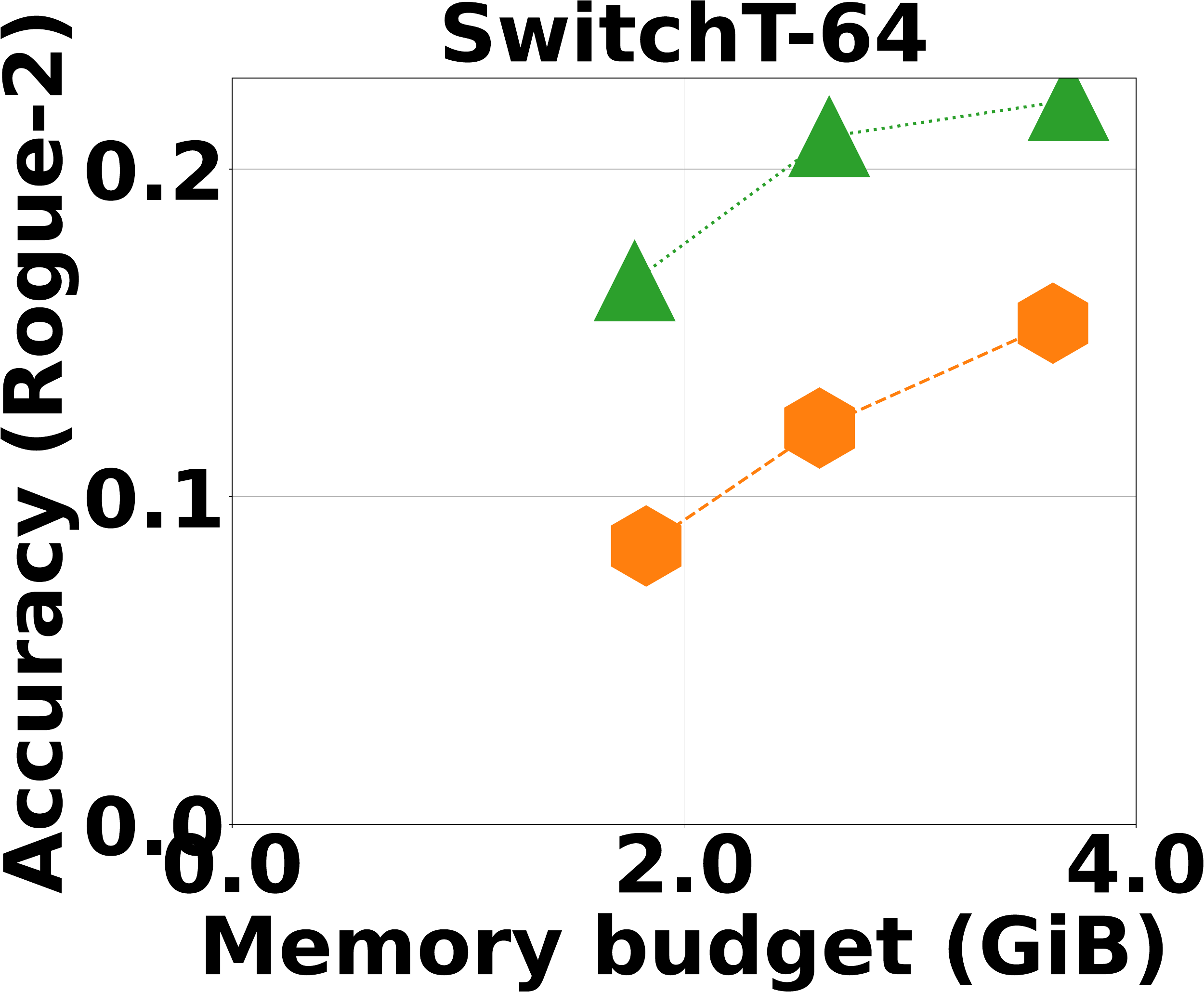}
        \end{subfigure}
        \hfill
        \begin{subfigure}[]{0.2\textwidth}\centering
            \includegraphics[width=0.99\textwidth]{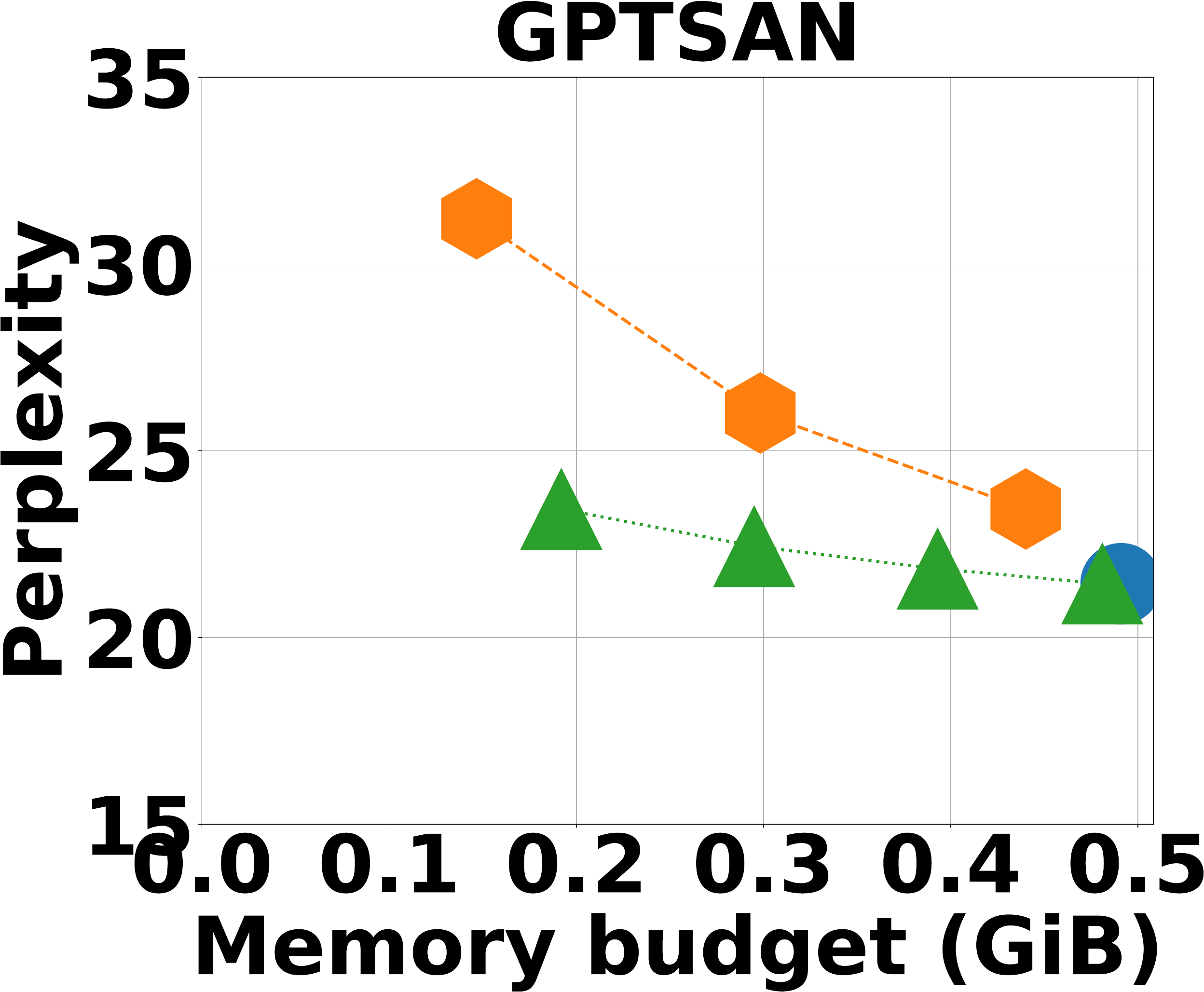}
        \end{subfigure}
        \caption{Jetson Nano}\label{result:memory-accuracy-tradeoff:nano}
    \end{subfigure}
    \vspace{-10pt}
    \caption{
        The accuracy achieved by \name and the pruning baseline under different memory budgets on (a) Jetson AGX and (b) Jetson Nano.
        \name achieves significantly higher accuracy than the baseline in almost all cases.
    }
    \vspace{-10pt}
\end{figure*}

\subsection{Overall Runtime Performance}

Our approach offers good memory-latency tradeoffs. As shown in Figure~\ref{result:memory-latency-tradeoff}, in terms of latency, our approach and the `Pruning' method exhibit very similar memory-latency trade-offs. This indicates that our approach, while ensuring minimal model accuracy degradation, enables the model to occupy less memory and achieve reduced inference latency. Although the `On demand' method can significantly reduce memory usage, its cost comes in the form of high latency overhead.

Our approach also offers good memory-accuracy tradeoffs.
As shown in Figure~\ref{result:memory-accuracy-tradeoff:agx} and Figure~\ref{result:memory-accuracy-tradeoff:nano}, \name creates a trade-off space that accommodates various resource usages. Higher resource utilization leads to better model performance and vice versa; and, even if resource consumption reduces, \name still maintains the model's performance. 
For example, in summarization task, \name reduces memory usage and latency by 37\% and 18\% with only 0.012 Rouge-2 degradation on Jetson AGX.

It is worth noting that SwitchT-64 cannot be directly deployed on Jetson Nano, as its hardware has a maximum memory support of 4GB, while the inference demand of SwitchT-16 exceeds 4GB. Nevertheless, in Figure~\ref{result:memory-accuracy-tradeoff:nano}, our approach significantly outperforms the 'Pruning' baseline. For instance, on Jetson Nano, when reducing the memory footprint of SwitchT-64 from 14GiB to 2.6GB, \name achieves an accuracy that is 175\% higher than the baseline.


\name also outperforms the baselines, demonstrating its effectiveness. Under the same memory budget, it achieves better performance than the baselines. 
For example, in SwitchT-32, when the memory usage is 4.7GB, the Rugue score of \name is 0.232, which is 8.76\% higher than the 'Pruning' baseline.
This is because we select the most important experts based on the characteristics of input samples, rather than simply pruning the MoE model into a smaller but static model. 
However, we cannot surpass the on-demand approach because it does not alter the model's output. However, its high latency makes it unsuitable for edge scenarios where both latency and memory are constrained.

\subsection{Offline Planning Performance}

\begin{table}
    \centering
    \resizebox{0.5\textwidth}{!}{
        \begin{tabular}{|c|c|ccc|}
            \hline
            \multirow{2}{*}{\textbf{Method}}   & \textbf{Constraints}         & \multicolumn{3}{c|}{\textbf{Achieved performance}}                                                        \\ \cline{2-5} 
                                               & \textbf{Memory budget (GiB)} & \multicolumn{1}{c|}{\textbf{Memory (GiB)}} & \multicolumn{1}{c|}{\textbf{Latency (s)}} & \textbf{Rogue-2} \\ \hline
            \textbf{Original MoE}              & \textbackslash{}             & \multicolumn{1}{c|}{4.08}                  & \multicolumn{1}{c|}{0.82}                 & 0.2             \\ \hline
            \multirow{4}{*}{\textbf{vExperts}} & 2.0                            & \multicolumn{1}{c|}{1.64}                  & \multicolumn{1}{c|}{0.45}                 & 0.21             \\ \cline{2-5} 
                                               & 3.0                            & \multicolumn{1}{c|}{2.74}                  & \multicolumn{1}{c|}{0.65}                 & 0.22             \\ \cline{2-5} 
                                               & 3.5                          & \multicolumn{1}{c|}{3.58}                  & \multicolumn{1}{c|}{0.76}                 & 0.23             \\ \cline{2-5} 
                                               & 4.0                            & \multicolumn{1}{c|}{3.99}                  & \multicolumn{1}{c|}{0.80}                 & 0.23             \\ \hline
            \end{tabular}
    }
    \caption{Performance of \name with SwitchT-16 under different memory constraints .}
    \label{table:profiled}
    \vspace{-10pt}
\end{table}



\name can find optimal configurations that satisfy given constraints while maximizing model performance, as shown in Table~\ref{table:profiled}. For example, when given a memory budget of 2.0 GiB, the configuration found by \name allows the model to achieve an actual 1.64 GiB peak GPU memory usage and 0.82 s inference latency, satisfying the constraints while experiencing a 0.01 Rogue-2 score increase compared to the original MoE model.

Furthermore, it is observed that as the constraints become looser, \name can achieve higher accuracy, and conversely, tighter constraints result in lower accuracy. 
It found optimal configurations under various constraints, respectively, to maximize the utilization of resources.
For example, in the summarization task, when the memory budget changed from 3.5 GiB to 4 GiB, the actual peak GPU memory and inference latency of the model during runtime changed from 3.58 GiB and 0.76 s to 3.99 GiB and 0.80 s, respectively. The accuracy also changed from 0.23 to 0.2.
However, our method is not always able to satisfy the budget. For example, in the summarization task with SwitchT-16, when the memory budget is set to 3.5 GiB, the actual peak memory usage and inference latency achieved by the model are 3.58 GiB and 0.76 s, respectively. 
This situation arises because the inference performance of the model itself is difficult to predict accurately~\cite{nnmeter:mobisys21:Zhang}, leading to inaccurate performance modeling.


\begin{figure}
    \centering
    \includegraphics[width=0.5\textwidth]{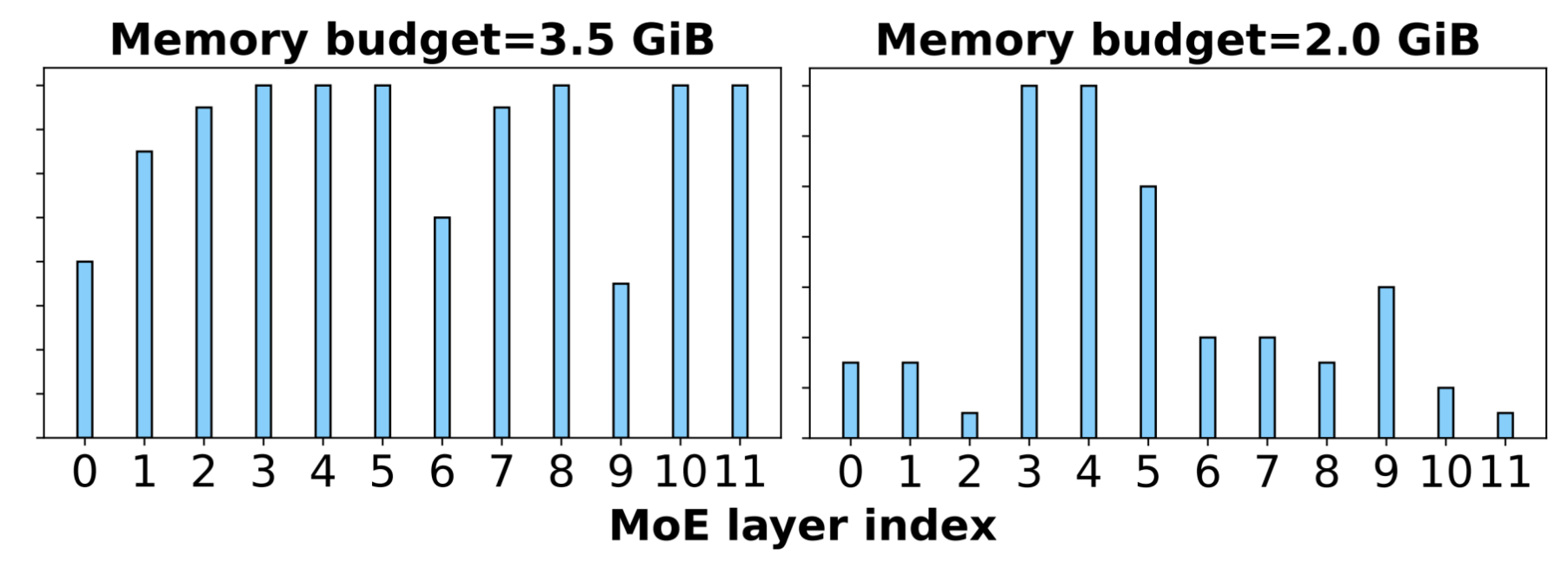}
    \vspace{-20pt}
    \caption{
        Distribution of allocated memory across layers under different memory budgets with SwitchT-16.
    }
    \label{result:distribution}
\end{figure}

Additionally, we find that the distribution of \core across different layers differs in different tasks. 
In the language modeling task, \name tends to maintain more experts in the middle layers, as shown in Figure~\ref{result:distribution} (a).
This suggests that for language models, the intermediate MoE layer has a greater impact on model inference, as the intermediate layers have a more significant influence on logical processing.

\subsection{Robustness Analysis}

\begin{figure}[]
    \centering
    \begin{minipage}[]{0.45\textwidth}
        \centering
        \begin{subfigure}[]{0.49\textwidth}\centering
            \includegraphics[width=\textwidth]{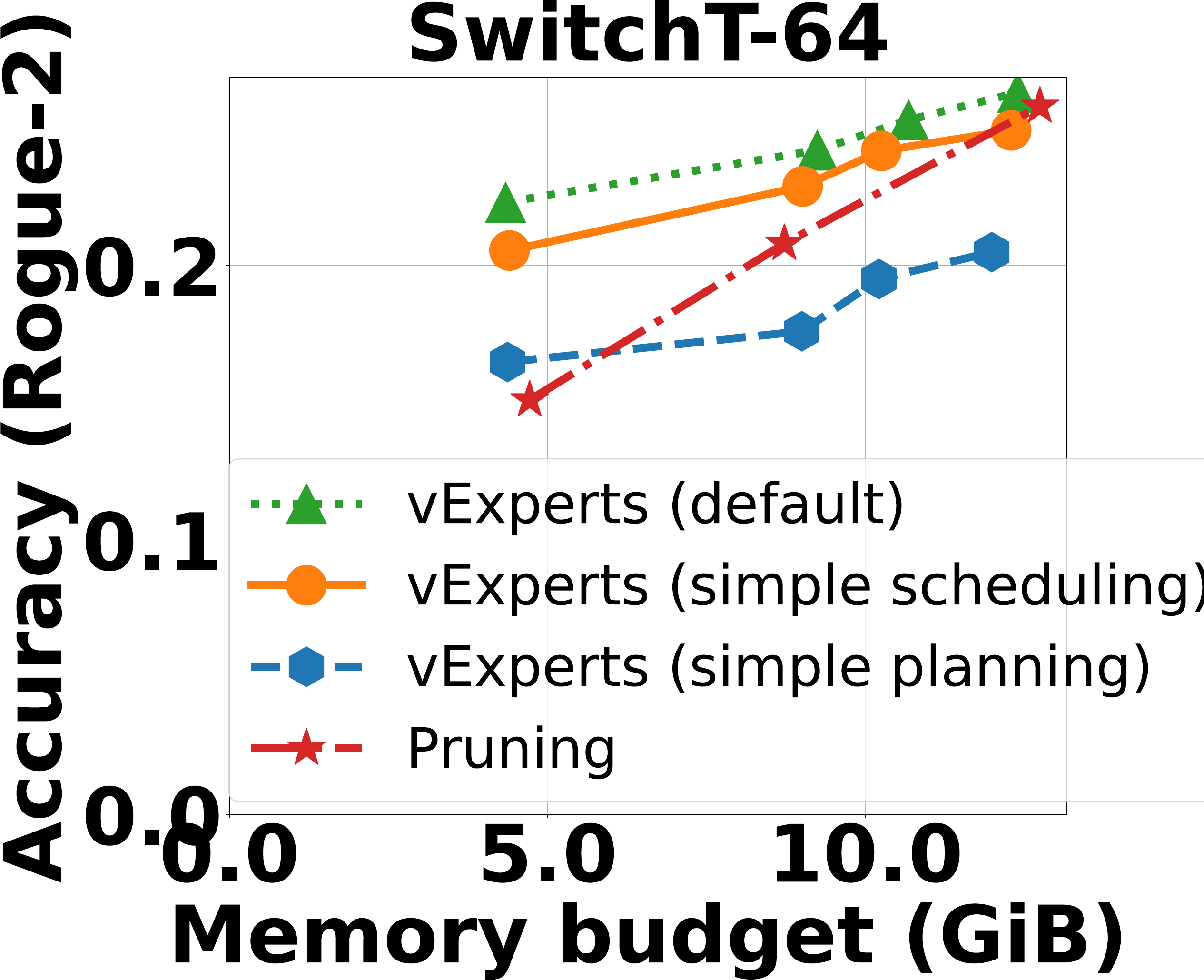}
        \end{subfigure}
        \begin{subfigure}[]{0.49\textwidth}\centering
            \includegraphics[width=\textwidth]{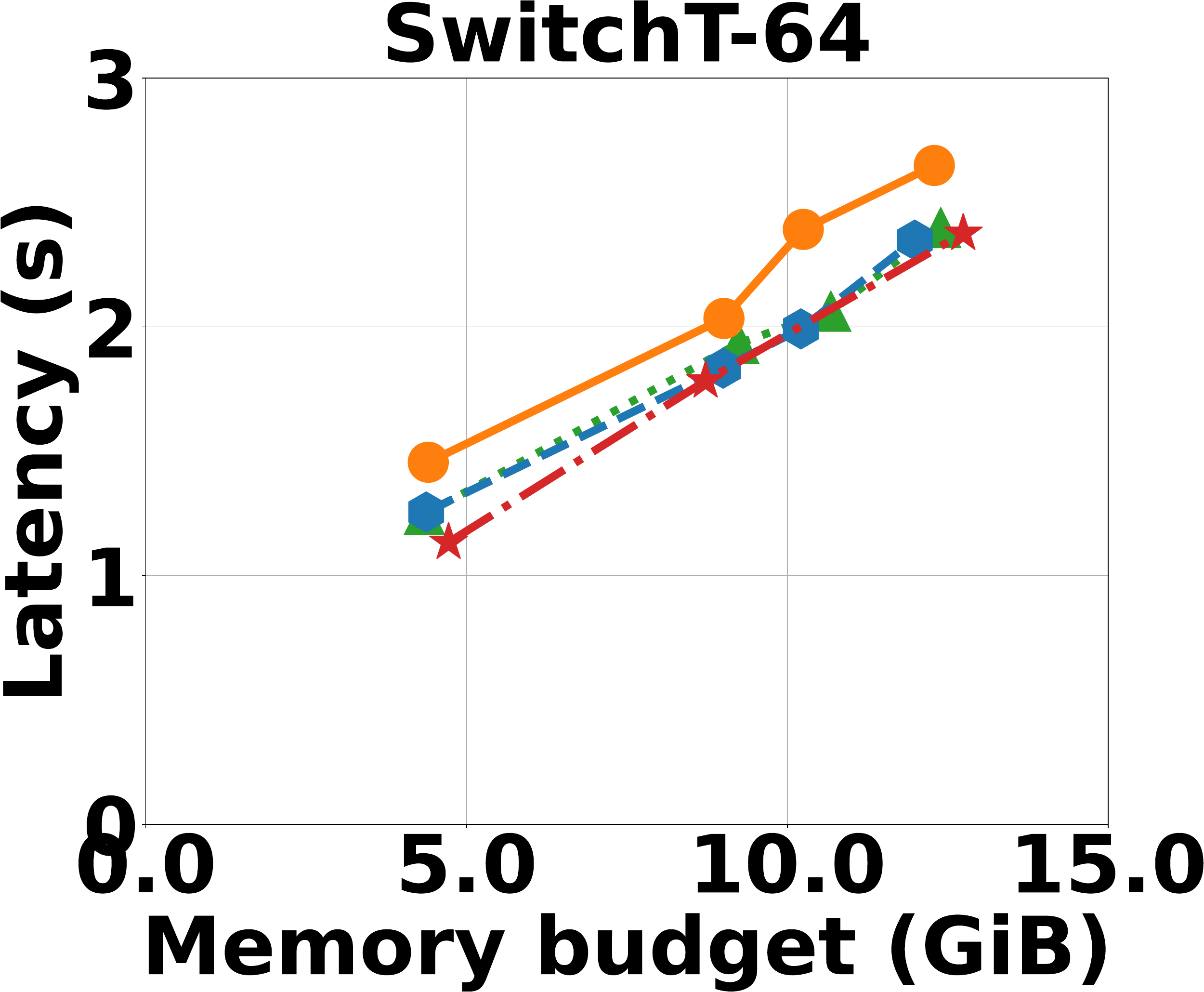}
        \end{subfigure}
        \vspace{-10pt}
        \caption{
            Ablation study: performance of \name with different components replaced.
        }
        \label{figure:ablation_study}
    \end{minipage}
    \vspace{-10pt}
\end{figure}

In this experiment, we analyze the robustness of \name across different usage scenarios.

\textbf{Different number of Experts.}
\name can be used for MoE models with different numbers of experts.
As shown in Figure~\ref{result:memory-accuracy-tradeoff:agx}, \name finds good performance-resource tradeoffs between Switch Transformers (16, 32, and 64) with different resource usage.
And \name can reduce the resource consumption of SwitchT-64 to a level similar to that of SwitchT-32, while maintaining comparable accuracy. This demonstrates that our method can effectively prune large MoE models into well-performing smaller models based on resource constraints, eliminating the need to store multiple model sizes.

\subsection{Ablation study}\label{sec:eval:breakdown}

In Figure~\ref{figure:ablation_study}, we show the performance of \name when replacing one component from the system. The experimental results indicate that the absence of any component in \name leads to performance degradation. In other words, all components contribute to the performance of \name.

(1) \textbf{`Simple scheduling'}: change the scheduling in \name component, remove amortized updating component, and calculate expert importance score where experts with more tokens to inference have higher scores. 
The perplexity of \name is better than \textbf{`simple scheduling'} because the importance score obtained by our method is more accurate via approximating the outputs of experts, instead of simplely counting the number of tokens dispatched to each expert.

(2) \textbf{`Simple planning'}: change the planning in \name component to \textbf{`simple planning'} and adopt an even \core distribution, meaning that each MoE layer has an equal number of \core, and then selects the best from these search spaces, where search space is very limited and does not encompass the optimal solution.
The perplexity of \name is superior to \textbf{`simple planning'} because the memory planner utilizes genetic search to find the optimal configuration for \name from a large configuration space. 
In contrast, \textbf{`simple planning'} cannot find a worse configuration than \name due to the suboptimal search process.

\begin{table}
    \centering
    \resizebox{0.46\textwidth}{!}{
        \begin{tabular}{|c|c|cc|}
            \hline
            \multicolumn{1}{|c|}{\textbf{Memory}}                               & \textbf{External Memory} & \multicolumn{2}{c|}{\textbf{IO Overhead (MiB/s)}} \\ 
            \multicolumn{1}{|c|}{\textbf{Constraint (GiB)}} &  \textbf{Consumed (GiB)}                                        & \textbf{Peak}  & \textbf{Mean}  \\ \hline
            \multicolumn{1}{|c|}{1.2}                                          & 1.02                                                 & 40         & 20              \\ \hline
            \multicolumn{1}{|c|}{1.5}                                          & 1.02                                                 & 36         & 27             \\ \hline
            \multicolumn{1}{|c|}{1.8}                                           & 1.02                                                 & 13         & 11              \\ \hline
            \end{tabular}
    }
    \caption{
        The runtime overhead of \name.
    }\vspace{-20pt}
    \label{table:overhead:online}
\end{table}

\subsection{Overhead}

We report the overhead of \name in the object detection task with the Swin-MoE model.

\textbf{Offline planning overhead.} 
The offline planning phase includes performance modeling and optimal configuration generation.
Performance modeling only needs to be done once for a MoE model, which takes about 20 minutes.
Optimal configuration generation needs to be done with each different resource constraint, which takes about 5 seconds.

\textbf{Runtime overhead.}
The runtime overhead includes calculating expert importance score, expert swapping, and different expert request handling strategies.
As shown in Table~\ref{table:overhead:online}, the peak IO overhead of \name is about 20 MiB/s, which is negligible compared to the IO bandwidth between the main memory and the external memory (e.g., 10-30 GiB/s for GPU-CPU over PCIe and 300-600 MiB/s for CPU-SSD).
This is because the design of \name enables us to swap only a small number of experts.
The ``External Memory Consumed'' means the space needed in the external memory (CPU memory or storage) to store the weights of experts.
We store the original MoE model parameters in external memory to reduce the usage in main memory.
\section{Related Work}


\textbf{Systems optimization for MoE model serving.}
Common techniques for optimizing MoE model serving include offloading and swapping memory. Huang et al. \cite{cachemoe:arxiv23:huang} propose to swap the experts from GPU memory to CPU memory to reduce the memory consumption of MoE models, incurring high latency overhead. SE-MoE \cite{SEMoE:arxiv22:Liang} utilizes Ring Memory offloading to reduce GPU memory consumption of MoE models. However, these methods are not suitable for resource-constrained devices with dynamic latency and memory constraints.

\textbf{Optimization for dynamic DL model serving.}
MoE is a type of dynamic neural networks. Many approaches are introduced to enable or enhance such dynamic DL model serving in general.
Nimble \cite{shen2021nimble} is a system that optimizes, compiles, and executes dynamic neural networks on multiple platforms by using a dynamic type system and a lightweight virtual machine runtime.
Model scaling approaches \cite{nestdnn:mobicom18:Fang,legodnn:mobicom21:Han,wen2023adaptivenet} propose to adjust the model size/architecture on edge devices to meet different resource constraints.
Remix \cite{jiang2021remix} proposes to use multiple models and dynamically switch between them during the inference process.
Besides serving dynamic models, researchers have also attempted to slice the static model to dynamic components to achieve different resource-performance tradeoffs \cite{neulens:mobicom22:Hou,zhang2020dynamic_slicing}.
As compared with these approaches, our design is fundamentally different because it is based on the unique structure and characteristics of MoE.

\textbf{Efficient design of MoE models.} 
Many existing approaches study the efficiency problem of MoE models from the model design perspective. GShard \cite{lepikhin2020gshard} scales up Transformer with MoE and improves the quality and efficiency of multilingual machine translation. 
Task-MoE \cite{moedistillation:arxiv21:Kudugunta} extract subnet from a large MoE model by a task-level routing strategy.
Chen et al. \cite{taskexpertpruning:arxiv22:Chen} propose to prune non-professional experts according to the downstream task for efficient MoE deployment. 
MPoE \cite{pmoe:arxiv22:Gao} proposes to build a parameter-efficient MoE architecture by enforcing parameter sharing between the experts. 
AutoMoE \cite{automoe:arxiv22:Jawahar} utilizes neural architecture search to automatically design more efficient MoE models. 
These methods need to modify and retrain the MoE model to reduce resource consumption. They are orthogonal to our approach and can use our method to further reduce resource consumption.
\section{Conclusion}

This paper addresses the challenges of deploying MoE models to resource-constrained edge devices. We propose a framework called \name that adaptively reduces the inference costs of MoE models based on the memory constraints while preserving the model accuracy. Experimental results have demonstrated that our method can significantly reduce the inference costs of MoE models with reasonable accuracy degradation.
\name creates a nice space of resource-accuracy trade-off of SoTA large MoE models. 
Our work does not have obvious ethical impacts, as we focusing on model inference acceleration.

\section{Limitations}

The tasks considered in this paper are relatively limited, and the proposed method has not been evaluated across a wide range of tasks. Subsequently, we will continue to expand our evaluation to other NLP tasks. The paper did not test larger MoE models (e.g., those exceeding 70B parameters) due to computational resource constraints. Instead, experiments were conducted only on smaller Switch Transformers models, demonstrating the effectiveness of the proposed method.

\section*{Acknowledgement}
\label{sec:acknowledgement}

This work is supported by the National Natural Science Foundation of China (Grant No.62272261), and is partly supported by Shuimu Tsinghua Scholar Program (Grant 2023SM201).

\bibliography{custom}

\appendix


\section{Details of Fine-grained Expert Profiling}
\subsection{Memory Footprint and Latency Estimation}
In order to estimate the MoE model's inference latency and memory, we need to conduct a more detailed profiling of the experts, encompassing the inference memory footprint, latency, and loading time. This is crucial as the overall cost is composed of numerous experts within the \core.

\textbf{Expert memory footprint}:
We conducted a detailed profiling of the memory footprint for the inference of each individual expert within the MoE layer. This includes the memory occupied by the parameters of each expert and the memory occupied by the activations generated during inference computation.
\textbf{Expert inference latency}:
The expert inference latency encompasses both data transmission and computation. We profiled the data transmission and computation time for each expert. Given that the parameters of each expert within a layer are identical, we only need to profile one expert per layer.
\textbf{Expert loading time}:
The loading time for each expert refers to the time taken for expert parameter transmission. As expert loading involves the transmission of data in I/O, we also need to profile the I/O resources to ensure that computation does not get blocked during expert loading.

With expert-level profiling data, we can obtain the whole model performance,
where the computational load during model inference is mainly related to the number of experts in the configuration: the more experts there are, the larger the inference latency and memory footprint will be, and vice versa.
While the expert's inference latency and memory differ across different hardware, we only need to model it once for a given hardware to obtain $E_\text{memory}$, and $E_\text{latency}$.


\subsection{Accuracy Influence Modeling}
Different distributions of \core across layers may lead to different influences to model accuracy. 
To obtain the optimal configuration of expert distribution, we must have the ability to efficiently obtain the accuracy of each configuration.
Directly measuring this accuracy influences on the target device is time-consuming because it requires running the MoE model under different configurations for multiple times.
Since the configuration space of expert distributions is large, measurement is impractical.
Therefore, we decide to use machine learning to model the accuracy influence based on profiling data.

Specifically, to obtain $E_\text{accuracy}$, we first need to collect a small amount of labeled profiling samples from the target device that can reflect the data distribution of the deployment scenario.
The samples are then used to measure the accuracy of \name under different configurations.
Since the deployed models will be used in the target environment for a long time, it is feasible to collect such profiling data.
The data labeling can be done manually or with an oracle model.

Next, we generate a set of random configurations of \core.
For each configuration, we use \name under the configuration to perform model inference with the profiling dataset.
We collect the corresponding $E_\text{accuracy}$. 
Note that when collecting $E_\text{accuracy}$, \name is operated by the runtime scheduler, which can refer to Section~\ref{sec:moe_inference} and Section~\ref{sec:expert_loading}.

Finally, we learn the relation between \core configurations and the model accuracy with a small DNN (containing two fully-connected layers with ReLU). The DNN is lightweight and sufficient. It minimizes the residual sum of squares between the actual accuracy and predicted $E_\text{accuracy}$.
The training of this DNN is exceptionally fast, with very low computational cost, typically converged in just a few minutes (on Jetson Nano) with the prediction error less than 1\%.

\end{document}